\documentclass{article}
\usepackage{iclr2027_conference,times}

\usepackage{amsmath,amssymb,amsthm}
\usepackage{booktabs,tabularx,array}
\usepackage{algorithm}
\usepackage{algpseudocode}
\usepackage{placeins}
\usepackage{flafter}
\usepackage{hyperref}
\usepackage{url}
\usepackage{microtype}
\usepackage{graphicx}
\usepackage{enumitem}
\DeclareRobustCommand{\methodname}{S\textsuperscript{3}-PnP}
\newtheorem{lemma}{Lemma}

\theoremstyle{definition}

\newtheorem{theorem}{Theorem}
\usepackage{wrapfig}
\usepackage{pifont}

\title{\methodname: Plug-and-Play Denoising for Scientific Imaging with Self-Supervised Score Priors}

\hypersetup{
    colorlinks=false,
    pdfborder={0 0 1},
    linkbordercolor={0 1 0},
    citebordercolor={0 1 0},
    urlbordercolor={0 1 0},
    filebordercolor={0 1 0},
    pdftitle={S³-PnP: Plug-and-Play Denoising for Scientific Imaging with Self-Supervised Score Priors}
}
\author{Yukun Cui$^{1}$\quad
Jiangshe Zhang$^{1}$\thanks{Corresponding authors.}\quad
Shuang Xu$^{2}$\quad
Lilun Deng$^{1}$\quad
Haowen Bai$^{3}$\\
\textbf{Tao Feng$^{4}$\quad
Haotong Qin$^{5}$\quad
Konrad Schindler$^{6}$\quad
Zixiang Zhao$^{6\,*}$}\\[1mm]
$^{1}$Xi’an Jiaotong University\quad
$^{2}$Northwestern Polytechnical University\\
$^{3}$Nanyang Technological University\quad
$^{4}$Tsinghua University\\
$^{5}$Hong Kong Polytechnic University\quad
$^{6}$ETH Z\"urich\\[1mm]
\texttt{cuiyukun@xjtu.edu.cn; zixiang.zhao@ethz.ch}
}

\iclrfinalcopy

\begin{document}

\maketitle
\lhead{Preprint}

\begin{abstract}
Modern image restoration methods increasingly rely on large-scale supervised training, yet clean targets are often prohibitively difficult to obtain in scientific imaging modalities such as hyperspectral imaging (HSI) and magnetic resonance imaging (MRI). Plug-and-play ADMM (PnP-ADMM) with pretrained diffusion priors offers a promising alternative, but the learned prior distribution can mismatch both the target scientific image distribution and the distribution of ADMM denoiser inputs induced by dual feedback. To mitigate these mismatches, we propose \mbox{\textbf{\methodname{}}}, a self-supervised plug-and-play restoration framework that learns a score prior directly from observation-derived ADMM intermediate states, without clean training targets or externally pretrained weights. Specifically, the learned score corrects the input to the analytical prior update, combining observation-driven corrections with analytical regularization. We further analyze the convergence and reconstruction error of the corrected ADMM algorithm assuming an ideal score and quadratic data fidelity. Experiments on reference-guided HSI restoration, single-image HSI denoising, and MRI denoising demonstrate consistent improvements over competing methods.
\end{abstract}

\section{Introduction}
\label{sec:introduction}

\begin{figure}[t]
    \centering
    \includegraphics[width=\linewidth]{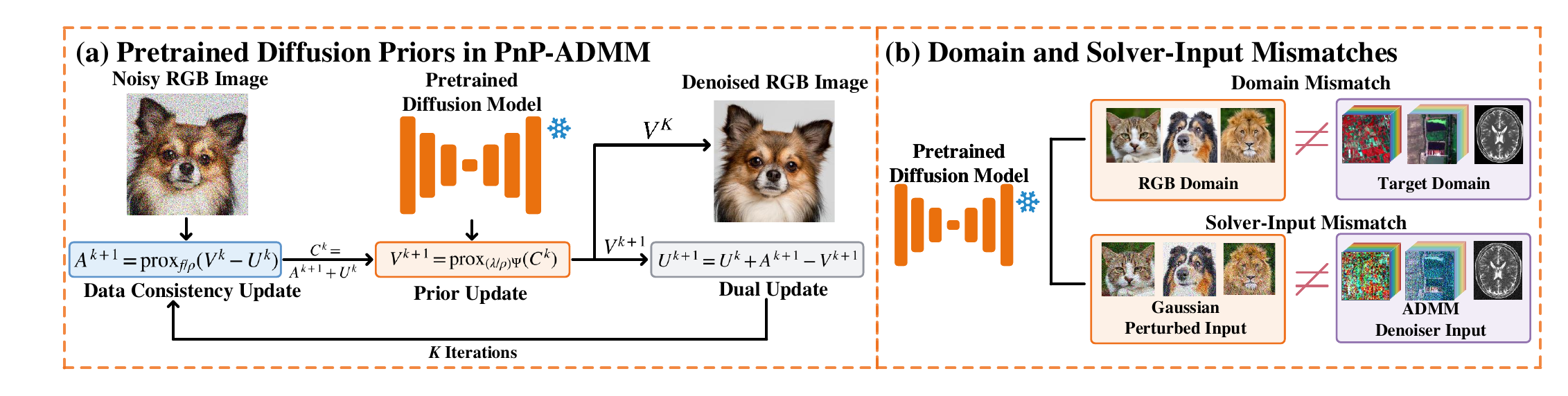}
    \caption{Two sources of mismatch when using pretrained image priors in scientific imaging: differences between the pretraining and target image domains, and differences between score-model training inputs and ADMM denoiser inputs.}
    \label{fig:motivation}
\end{figure}

Recovering high-quality images from noisy measurements is a fundamental challenge in scientific imaging~\citep{zhu2018image}. In modalities such as hyperspectral imaging (HSI) and magnetic resonance imaging (MRI), noise degrades image quality and obscures important structures. However, clean reference images are often difficult to acquire, limiting the availability of paired noisy and clean data for supervised end-to-end restoration. The alternating direction method of multipliers (ADMM) provides a model-based framework that combines observation fidelity with analytical priors~\citep{boyd2011admm}. To describe the ADMM formulation, let $A$ denote the image and $f(A)$ the data-fidelity term. Introducing an auxiliary variable $V$ yields:
\begin{equation}
    \min_{A,V}\ f(A)+\lambda\Psi(V)
    \qquad\text{subject to}\qquad A=V.
    \label{eq:split-objective}
\end{equation}
Here $\Psi$ is an analytical regularizer and $\lambda>0$ controls its strength. With a scaled dual variable $U$ and penalty parameter $\rho>0$, ADMM performs the following updates:
\begin{equation}
    \begin{aligned}
        A^{k+1}=\operatorname{prox}_{(f/\rho)}(V^k-U^k)&=\arg\min_A f(A)+\frac{\rho}{2}\|A-V^k+U^k\|_2^2,\\
        C^k=A^{k+1}+U^k,\qquad
        V^{k+1}&=\mathcal T_k(C^k),\qquad
        U^{k+1}=U^k+A^{k+1}-V^{k+1}.
    \end{aligned}
    \label{eq:admm}
\end{equation}
In these updates, $C^k$ denotes the \emph{ADMM denoiser input}, combining the data update with dual feedback, and $\mathcal T_k$ denotes the analytical prior operator. ADMM-based PnP methods replace this analytical prior operator with a denoising operator $\mathcal C_k$, yielding $V^{k+1}=\mathcal C_k(C^k)$ \citep{venkatakrishnan2013pnp}.

Diffusion-based PnP methods and ADMM variants often use diffusion models pretrained on natural RGB images \citep{zhu2023diffpir,wu2024pnpdm,zhang2025admmdiff,shrestha2026taming}. Figure~\ref{fig:motivation}(a) illustrates how a pretrained diffusion model supplies the prior update within ADMM. Applying such priors to scientific imaging can introduce two distinct mismatches, illustrated in Figure~\ref{fig:motivation}(b).
First, natural RGB images differ from scientific images in their image statistics and representations, creating a mismatch between the pretraining and target domains. Adapting these priors to modalities such as hyperspectral imaging and MRI may therefore require additional representation or model adaptation \citep{pang2024hirdiff,xiao2025hspdiff,gao2024mrpd}.
Second, even within the same modality, ADMM denoiser inputs need not follow the Gaussian perturbation distributions used to train score models, because they also reflect dual feedback and solver-dependent errors. AC--DC mitigates this mismatch by adapting solver inputs to a pretrained score model through noise injection and conditional Langevin correction \citep{shrestha2026taming}. This motivates a complementary direction: learning score corrections directly from observation-derived ADMM states, so that their statistics inform the prior update without requiring an externally pretrained model.

To address these mismatches in scientific image restoration, we introduce \methodname{} for settings without domain-matched pretrained priors. The framework learns a self-supervised score approximation at test time from ADMM denoiser inputs generated by the available observations. The learned score corrects the input to the analytical prior update, combining observation-adapted state statistics with analytical regularization. Training requires neither clean targets nor externally pretrained weights. Under a matched quadratic denoising model with ideal score corrections, we establish sufficient conditions for convergence of the corrected ADMM iterations and lower expected reconstruction error relative to the matched analytical ADMM baseline. Because structured interference may enter the learned state statistics, a structure-guided learning strategy combines complementary orthogonal projections with sequential training. We evaluate \methodname{} on reference-guided HSI restoration, single-image HSI denoising, and MRI denoising.
Our contributions are:

\begin{itemize}[itemsep=0pt, topsep=0pt, parsep=0pt]

\item \textbf{Self-supervised score prior learning.}
For scientific image restoration, \methodname{} learns score corrections from ADMM states at test time to modify the analytical prior input.
Training requires neither clean targets nor externally pretrained weights.

\item \textbf{Convergence and reconstruction error analysis.}
We establish sufficient conditions for \methodname{} to converge and achieve lower expected reconstruction error than matched analytical ADMM, assuming an ideal score and quadratic data fidelity.

\item \textbf{Score learning in complementary orthogonal subspaces.}
We learn and independently control score corrections in complementary orthogonal subspaces, enabling separate treatment of corrections associated with structured interference.
\end{itemize}
\section{Related Work}
\label{sec:related_work}

\noindent\textbf{Plug-and-Play and Generative Priors for Image Restoration.} 
\label{sec:related-pnp}
PnP combines denoisers with observation-model updates \citep{venkatakrishnan2013pnp}, with subsequent analyses covering fixed-point convergence, Lipschitz conditions, and proximal denoiser constructions \citep{chan2016pnp,ryu2019pnpconvergence,hurault2022proximal}. For RGB restoration, DDRM uses spectral decompositions of linear observation operators \citep{kawar2022ddrm}, while DDNM uses range--null-space decomposition to enforce data consistency \citep{wang2023ddnm}.
DPS and $\Pi$GDM guide diffusion sampling through approximate measurement likelihoods \citep{chung2023dps,song2023pigdm}.
DiffPIR combines diffusion denoising with half-quadratic splitting \citep{zhu2023diffpir}, and PnP-DM uses split Gibbs sampling for likelihood and prior updates \citep{wu2024pnpdm}.
PnP-Flow constructs time-dependent denoisers from pretrained flow-matching models \citep{martin2025pnpflow}, while CM4IR uses consistency models for few-step restoration \citep{garber2025cm4ir}.
ADMMDiff coordinates generative priors and guidance within ADMM \citep{zhang2025admmdiff}. For non-RGB imaging, PLRDiff and HIR-Diff use low-dimensional spectral representations to apply pretrained three-channel diffusion priors to HSI restoration \citep{rui2024plrdiff,pang2024hirdiff}.
DM-ZS learns guidance and spectral mapping from PAN and low-resolution HSI observations for pansharpening with a pretrained RGB diffusion model \citep{xiao2025hspdiff}.
For MRI reconstruction, \citet{chung2022scoremri} combine an MRI-trained score prior with data-consistency updates, while DDS couples diffusion sampling with conjugate-gradient updates \citep{chung2024dds}.
MRPD transfers a natural-image latent diffusion prior to MRI with optional autoencoder adaptation \citep{gao2024mrpd}, and PnP-CM incorporates MRI-trained consistency models into PnP-ADMM \citep{gulle2026pnpcm}. Our framework combines learned score corrections with an analytical prior update within ADMM.
The corrections act on the prior input, allowing learned information to complement analytical regularization.

\noindent\textbf{Self-Supervised Prior Learning and Adaptation.} 
\label{sec:related-score}
Self-supervised learning and prior adaptation address limited clean supervision and mismatch with the target restoration problem.
\citet{shoushtari2024mismatch} analyze denoiser mismatch in PnP-ADMM and investigate adaptation using target-domain samples.
AC--DC addresses the mismatch between pretrained score models and ADMM inputs shaped by dual feedback through Gaussian noise injection and conditional Langevin corrections \citep{shrestha2026taming}.
To learn without clean targets, Noise2Self constructs self-supervised losses using noise independence and masked prediction \citep{batson2019noise2self}.
For score estimation, DSM connects perturbation-based denoising regression to distribution scores \citep{vincent2011dsm}.
Noise2Score learns the score of the noisy-image distribution and uses Tweedie's formula for restoration \citep{kim2021noise2score}, while C2S introduces generalized denoising score matching on further-corrupted observations for MRI denoising \citep{tu2025corruption2self}.
Learning directly from the observations to be restored provides another route to reducing external data requirements.
ZS-N2N trains a lightweight network on downsampled pairs from a single noisy image \citep{mansour2023zsn2n}, and DDS2M optimizes spatial and spectral networks during reverse diffusion using only the degraded HSI \citep{miao2023dds2m}.
For diffusion MRI, DDM\textsuperscript{2} combines statistical denoising with generative diffusion in a three-stage framework \citep{xiang2023ddm2}; Di-Fusion uses later diffusion steps and adaptive sampling in a single-stage self-supervised framework \citep{wu2025difusion}.
We learn score corrections at test time from ADMM states generated by the available observations.
These corrections modify the input to an analytical prior update, combining observation-driven learning with analytical regularization without externally pretrained weights.

\section{Method}
\label{sec:method}

\subsection{Self-Supervised Score Approximation from ADMM Iterates}
\label{sec:method-score}

Analytical priors in ADMM capture structures such as low rank and sparsity, but may not fully describe the statistics of the denoiser inputs generated during optimization. These statistics can complement analytical regularization, helping suppress noise while preserving image content. Therefore, \methodname{} learns an implicit regularization correction from ADMM denoiser inputs generated by the current observations. We estimate this correction through self-supervised score regression without externally pretrained weights and apply it before the analytical prior update.

\noindent\textbf{Self-supervised score learning.}
Firstly, we run Eq.~\eqref{eq:admm} with its analytical prior update and collect the denoiser inputs $C^k=A^{k+1}+U^k$ into a training set $\mathcal B$ to capture observation-dependent state statistics. Each state $C\in\mathcal B$ is normalized as $Z=\mathcal S(C)=(C-m)/ s$, using fixed medians $m$ and positive scales $s$ estimated from $\mathcal B$. Then, we construct the perturbed state $Z_t=Z+\sqrt{\Delta_t}\epsilon$, where $\epsilon\sim\mathcal N(0,I)$ is independent of $Z$ and $\Delta_t>0$ is the artificial perturbation variance.
In order to formulate a correction direction from these state statistics, let $p_t$ denote the density of $Z_t$ and define the state energy $R_t(z)=-\log p_t(z)$. The score of the perturbed state distribution is its log-density gradient, $\nabla_z\log p_t(z)=-\nabla_zR_t(z)$, which provides a correction direction reflecting the reference-state statistics. Since $p_t$ is unknown, we estimate this score through self-supervised regression using the collected states and their artificial perturbations. Specifically, we build on the regression principle of generalized denoising score matching (GDSM) \citep{tu2025corruption2self}. The following lemma establishes the relationship between its optimal restoration predictor and the state score, providing the basis for learning the required correction without clean training targets.

\begin{samepage}
\begin{lemma}
\label{lem:state-gdsm}
Under the state construction and fixed normalization above, assume $\mathbb E\|Z\|^2<\infty$. For fixed $t$ and $\kappa_t>\Delta_t>0$, consider the GDSM interpolation loss
\begin{equation}
    \mathcal L_{\mathrm{GDSM}}(\widehat Z;t)
    =
    \mathbb E\!\left\|
        \frac{\Delta_t}{\kappa_t}\widehat Z(Z_t,t)
        +\left(1-\frac{\Delta_t}{\kappa_t}\right)Z_t-Z
    \right\|^2.
    \label{eq:gdsm-interpolation-loss}
\end{equation}
Its minimizer over square-integrable restoration predictors satisfies, almost surely,
\begin{equation}
    \widehat Z^\star(Z_t,t)
    =
    Z_t+\kappa_t s_t^\star(Z_t),
    \label{eq:tweedie-score}
\end{equation}
where $s_t^\star=\nabla\log p_t=-\nabla R_t$. Moreover, $\mathbb E[Z\mid Z_t]=Z_t+\Delta_t s_t^\star(Z_t)$. If additionally $Z=Z_0+\sigma_0\epsilon_0$, where $Z_0$ denotes the underlying clean state in normalized coordinates, $\sigma_0>0$, $\epsilon_0\sim\mathcal N(0,I)$, and $Z_0,\epsilon_0,\epsilon$ are mutually independent, choosing $\kappa_t=\sigma_0^2+\Delta_t$ gives $\widehat Z^\star(Z_t,t)=\mathbb E[Z_0\mid Z_t]$.
\end{lemma}
\end{samepage}
See Appendix~\ref{app:proof-state-score} for the proof.

To learn the state score directly, we parameterize the restoration predictor as $\widehat Z_\theta(Z_t,t)=Z_t+\kappa_t s_\theta(Z_t,t)$ in Eq.(\ref{eq:tweedie-score}), where $\theta$ denotes the trainable parameters of the correction network. When training a shared network across perturbation levels, differences in regression-error scale can affect their relative contributions to optimization. We therefore use noise-level-dependent normalization of the scaled correction error and express the resulting objective as weighted score regression. For fixed $t$, before task-specific output processing, the GDSM interpolation loss satisfies
\begin{equation}
    \begin{aligned}
        \mathcal L_{\mathrm{GDSM}}(\widehat Z_\theta;t)
        &=
        \frac{\Delta_t^2}{w(t)}
        \mathcal L_{\mathrm{score}}(\theta;t)=
        \frac{\Delta_t^2}{w(t)}
        \mathbb E\!\left[
            w(t)\left\|
                s_\theta(Z_t,t)
                -\Delta_t^{-1}(Z-Z_t)
            \right\|^2
        \right].
    \end{aligned}
    \label{eq:gdsm-score-loss}
\end{equation}
Here $w(t)=\kappa_t^2(N\gamma\sigma_t^2)^{-1}$, where $N$ is the number of tensor entries per sample, $\gamma>0$ controls the loss scale, and $\sigma_t>1$ specifies the perturbation level. For each fixed $t$, the factor $\Delta_t^2/w(t)$ is positive and independent of $\theta$, so the two objectives share the same minimizer under this parameterization. Training uses $\mathcal L_{\mathrm{score}}$ averaged over the sampled perturbation levels, with $w(t)$ determining their relative emphasis.

We then use the learned score to correct the input to the analytical prior update. With $Z=\mathcal S(C)$, the corrected operator is
\begin{equation}
    \mathcal C_k(C)
    =\mathcal T_k\!\left(
        \mathcal S^{-1}\!\left[
            Z+\beta_k\kappa_{t_k}s_\theta(Z,t_k)
        \right]\right).
    \label{eq:g-interface}
\end{equation}
Here $\beta_k\geq0$ controls the correction strength, $\kappa_{t_k}$ is the score-scaling coefficient introduced in Lemma~\ref{lem:state-gdsm}, $t_k$ specifies the smoothing level, and $\mathcal T_k$ is the analytical prior update.
For the ideal score, $s_{t_k}^\star=-\nabla R_{t_k}$, so the correction is a gradient step on the state energy in normalized coordinates.
Setting $\beta_k=0$ recovers the analytical ADMM solver in Eq.~\eqref{eq:admm}. We refer to the resulting ADMM framework with self-supervised score corrections as \methodname{}.

\noindent\textbf{Convergence and reconstruction error.} Next, we establish sufficient conditions under which the corrected ADMM algorithm converges and its limiting estimate has lower expected reconstruction error than that of analytical ADMM.
The analysis assumes a matched statistical model with an ideal score, fixed operators, and positive-definite quadratic fidelity.
\begin{theorem}
\label{prop:gdsm-recovery}
Under the clean-state and moment assumptions of Lemma~\ref{lem:state-gdsm}, let $A_0=\mathcal S^{-1}(Z_0)$, $Y=\mathcal S^{-1}(Z_t)$. Consider ADMM with fixed $\rho>0$, $H=H^\ast\succ0$, and $f(A)=\frac12\langle A-Y,H(A-Y)\rangle$. For $\beta_k\equiv\beta\in[0,1]$, use $V^{k+1}=\mathcal T(C^k+\beta d_t(C^k))$, where the ideal correction $d_t(C)=\kappa_t s\odot s_t^\star(\mathcal S(C))$ is $L$-Lipschitz.
Set $r_H=\max_j|(\rho-\lambda_j(H))/(\rho+\lambda_j(H))|$, where $\lambda_j(H)>0$ are the eigenvalues of $H$.

\textnormal{(i) \textbf{Convergence}.}
Let $C_\beta^k=A_\beta^{k+1}+U_\beta^k$ denote the ADMM denoiser input at iteration $k$ with correction strength $\beta$.
If $r_H(1+2\beta L)<1$, the iterates converge from any finite initialization and satisfy
\begin{equation}
    \|C_\beta^k-C_\beta^\star\|
    \le q_\beta^k\|C_\beta^0-C_\beta^\star\|,
    \label{eq:gdsm-main-convergence}
\end{equation}
where $C_\beta^\star$ is the unique fixed point of the denoiser-input iteration and $q_\beta=\tfrac12[1+r_H(1+2\beta L)]<1$.
Denote the common limit of the primal iterates $A_\beta^k$ and $V_\beta^k$ by $A_\beta^\star$.

\textnormal{(ii) \textbf{Error improvement}.}
Under the same stability condition, additionally assume $\|\mathcal T(Q)-Q\|\le b$ for all $Q$, with $b\ge0$.
Let $a_t=(\mathbb E\|Y-\mathbb E[A_0\mid Y]\|^2)^{1/2}$, with $\delta_\beta$, $c_\beta$, and $b_H$ defined in Appendix~\ref{app:proof-gdsm-recovery}.
For $\beta>0$, if $\delta_\beta<1$ and $c_\beta a_t+b_H/(1-\delta_\beta)<a_t-b_H$, then
\begin{equation}
    \mathbb E\|A_\beta^\star-A_0\|^2
    <\mathbb E\|A_{\mathrm{off}}^\star-A_0\|^2.
    \label{eq:gdsm-risk-improvement}
\end{equation}
Here $A_{\mathrm{off}}^\star$ is the converged estimate of the analytical ADMM solver in Eq.~\eqref{eq:admm}, recovered by setting $\beta=0$ while keeping the data term, analytical map, and penalty unchanged.
\end{theorem}
Appendix~\ref{app:proof-gdsm-recovery} gives the remaining constants and proof. We further examine the theoretical predictions on convergence and reconstruction error through controlled experiments in Section~\ref{sec:theory-experiments} and Appendix~\ref{app:theory-experiment-details}.
In practical imaging, structured interference may persist in the reference states and be reflected in the learned score.
Therefore, we introduce structure-guided score corrections in Section~\ref{sec:method-orthogonal}.

\subsection{Self-Supervised Score Learning and \methodname\ Inference}
\label{sec:method-orthogonal}
\label{sec:method-training}

\begin{figure}[t]
    \centering
    \includegraphics[width=\linewidth]{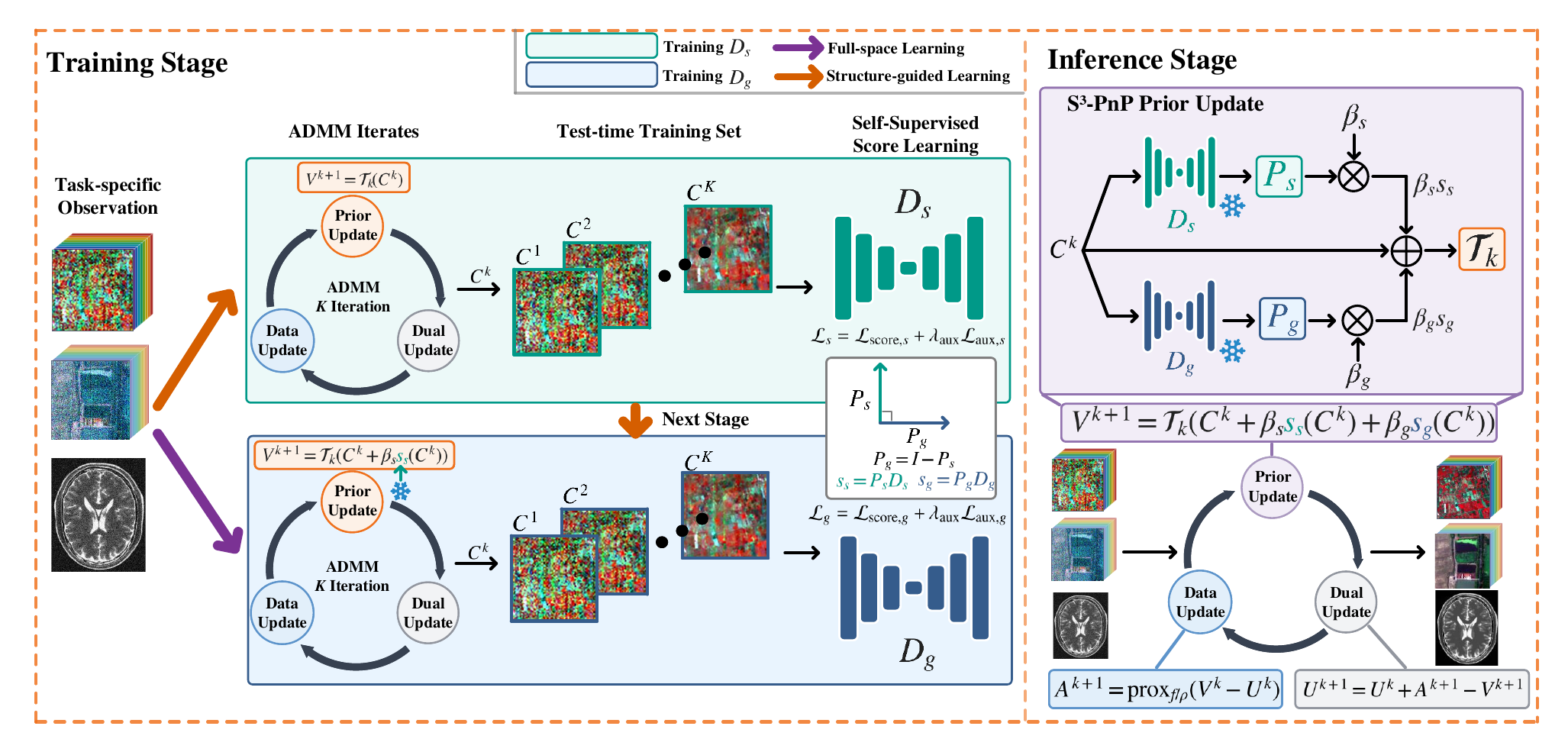}
    \caption{Training and inference of \methodname{}.
    Purple arrows indicate full-space learning with $P_s=0$, $P_g=I$, and $\beta_s=0$.
    Orange arrows indicate structure-guided learning: $D_s$ is trained and frozen, and its correction is enabled to collect a new state training set for $D_g$.
    At inference, active corrections are evaluated at the same ADMM input and combined before a single analytical prior update.
    Snowflakes denote frozen networks.}
    \label{fig:framework}
\end{figure}

The previous subsection described how the state score corrects the input to the analytical prior update.
Here, we explain how to learn these corrections, following the two procedures in Figure~\ref{fig:framework}: full-space learning and structure-guided learning.
Full-space learning estimates corrections over the entire space, while structure-guided learning separately controls correction components associated with structured interference.
For full-space learning, we use the lower module with $P_s=0$, $P_g=I$, and $\beta_s=0$.
Training begins by running analytical ADMM without learned corrections to collect denoiser inputs $C^k=A^{k+1}+U^k$ into a state training set $\mathcal B_0$.
We then use these states and their artificial perturbations to train the full-space network $D_g$ through self-supervision, as summarized in Algorithm~\ref{alg:score-learning}.
After training, we freeze the network and add its correction to the current state before the analytical prior update during inference.

However, structured interference, such as column stripes in hyperspectral images, can remain in the ADMM states used for full-space learning and influence the learned score.
To control stripe-related corrections separately, we use imaging knowledge to construct a subspace associated with column stripes and its orthogonal complement.
These subspaces are represented by orthogonal projectors $P_s$ and $P_g=I-P_s$, satisfying $P_s+P_g=I$ and $P_sP_g=0$.
The two correction components are then learned separately, with their strengths adjusted independently.
For a fixed reference-state distribution, let $s^\star$ denote the ideal score-induced correction and define $s_i^\star=P_i s^\star$ for $i\in\{s,g\}$.
The correction then decomposes as
\begin{equation}
    s^\star
    =P_s s^\star+P_g s^\star
    =s_s^\star+s_g^\star.
    \label{eq:projected-tweedie}
\end{equation}
The projector construction is detailed in Appendix~\ref{app:projector-construction}.
For structure-guided learning, we use two lightweight conditional U-Nets, $D_s$ and $D_g$, with the same architecture and independent parameters $\theta_s$ and $\theta_g$.
Following the orange path in Figure~\ref{fig:framework}, we first train the structure network $D_s$ on the state training set $\mathcal B_0$ generated by analytical ADMM.
We then freeze this network and rerun ADMM with its correction enabled to collect a new state training set $\mathcal B_1$ for training the complementary network $D_g$.
The two stages use their respective fixed normalizations and reference-state distributions.
We train each network using the self-supervised objective
\begin{equation}
    \mathcal L_i
    =\mathcal L_{\mathrm{score},i}
    +\lambda_{\mathrm{aux}}\mathcal L_{\mathrm{aux},i},
    \qquad i\in\{s,g\},
    \label{eq:branch-training-loss}
\end{equation}
where the score-regression term follows Eq.~\eqref{eq:gdsm-score-loss}, with task-specific output processing and auxiliary losses detailed in Appendix~\ref{app:training-details}.
During inference, the networks, normalizations, and deployment projectors remain fixed.
The active networks process the same ADMM input $C^k$.
Let $s_i(C^k,t_k;\theta_i)$ denote the learned correction in the original coordinates after rescaling, output processing, and projection by $P_i$. The score-scaling factor $\kappa_{t_k}$ is included in $s_i$.
When $\beta_i=0$, branch $i$ is skipped and its correction is set to zero.
As illustrated on the right of Figure~\ref{fig:framework}, these corrections are combined before a single analytical prior update:
\begin{equation}
    V^{k+1}
    =\mathcal T_k\!\left(
        C^k+\beta_s s_s(C^k,t_k;\theta_s)
        +\beta_g s_g(C^k,t_k;\theta_g)
    \right).
    \label{eq:orth-deployment}
\end{equation}
Here $\beta_s,\beta_g\geq0$ control the correction strengths, while the data and dual updates follow Eq.~\eqref{eq:admm}.
Setting both gains to zero recovers the matched analytical ADMM solver.
Algorithms~\ref{alg:sequential-training} and~\ref{alg:inference} summarize sequential training and unified inference, respectively.

\newsavebox{\SthreeLearnBox}
\newsavebox{\SthreeSequentialBox}
\newsavebox{\SthreeInferenceBox}
\newsavebox{\SthreeMathBox}
\newlength{\SthreeColumnWidth}
\newlength{\SthreeLeftGap}
\newlength{\SthreeMathWidth}
\makeatletter
\newcommand{\SthreeMath}[1]{%
    \begingroup
    \setlength{\SthreeMathWidth}{\dimexpr\linewidth-\ALG@tlm\relax}%
    \sbox{\SthreeMathBox}{$#1$}%
    \ifdim\wd\SthreeMathBox>\SthreeMathWidth
        \resizebox{\SthreeMathWidth}{!}{\usebox{\SthreeMathBox}}%
    \else
        \usebox{\SthreeMathBox}%
    \fi
    \endgroup
}
\makeatother

\begin{figure}[!htbp]
    \centering
    \setlength{\intextsep}{0pt}
    \setlength{\SthreeColumnWidth}{0.485\linewidth}
    \algrenewcommand{\algorithmicindent}{0.9em}
    \algrenewcommand{\algorithmicrequire}{\textbf{Input:}}
    \algrenewcommand{\algorithmiccomment}[1]{\hfill{\footnotesize$\triangleright$~#1}}
    \sbox{\SthreeLearnBox}{%
        \begin{minipage}{\SthreeColumnWidth}
            \begin{algorithm}[H]
                \caption{\textsc{LearnScore}$(\mathcal B,i)$}
                \label{alg:score-learning}
                \small
                \raggedright
                \begin{algorithmic}[1]
                    \Require State training set $\mathcal B$, branch $i$, $M$
                    \State Initialize $\theta_i$
                    \For{$n=1,\ldots,M$}
                        \State Sample $Z$ from $\mathcal B$, $t$, $\epsilon\sim\mathcal N(0,I)$
                        \State \SthreeMath{Z_t\gets Z+\sqrt{\Delta_t}\epsilon}
                        \State \SthreeMath{w(t)\gets\kappa_t^2(N\gamma\sigma_t^2)^{-1}}
                        \State \SthreeMath{\mathcal L_i\gets\mathcal L_{\mathrm{score},i}+\lambda_{\mathrm{aux}}\mathcal L_{\mathrm{aux},i}}
                        \State \SthreeMath{\theta_i\gets\operatorname{Adam}\!\left(\theta_i,\nabla_{\theta_i}\mathcal L_i\right)}
                    \EndFor
                    \State \Return $\theta_i$
                \end{algorithmic}
            \end{algorithm}
        \end{minipage}%
    }
    \sbox{\SthreeSequentialBox}{%
        \begin{minipage}{\SthreeColumnWidth}
            \begin{algorithm}[H]
                \caption{Structure-guided Training}
                \label{alg:sequential-training}
                \label{alg:dual-training}
                \small
                \raggedright
                \begin{algorithmic}[1]
                    \Require Visible observations, $P_s$, $P_g=I-P_s$, $\beta_s>0$
                    \State \SthreeMath{\text{Run ADMM via Eq.~}\eqref{eq:orth-deployment},\qquad\beta_s=\beta_g=0}
                    \State \SthreeMath{\mathcal B_0\gets\{C^k\}_{k\in\mathcal K}}
                    \State \SthreeMath{\theta_s\gets\text{\textsc{LearnScore}}(\mathcal B_0,s)} \Comment{Alg.~\ref{alg:score-learning}}
                    \State Freeze $\theta_s$
                    \State \SthreeMath{\text{Rerun ADMM via Eq.~}\eqref{eq:orth-deployment}\text{ with }\theta_s,\qquad\beta_g=0}
                    \State \SthreeMath{\mathcal B_1\gets\{C^k\}_{k\in\mathcal K}}
                    \State \SthreeMath{\theta_g\gets\text{\textsc{LearnScore}}(\mathcal B_1,g)} \Comment{Alg.~\ref{alg:score-learning}}
                    \State \Return $(\theta_s,\theta_g)$
                \end{algorithmic}
            \end{algorithm}
        \end{minipage}%
    }
    \sbox{\SthreeInferenceBox}{%
        \begin{minipage}{\SthreeColumnWidth}
            \begin{algorithm}[H]
                \caption{\methodname{} Inference}
                \label{alg:inference}
                \label{alg:admm-corrections}
                \small
                \raggedright
                \begin{algorithmic}[1]
                    \Require Noisy observations, $\{\theta_i,\beta_i\}_{i\in\{s,g\}}$, $P_s$, $\rho$, $K$, $\{t_k\}$
                    \State Initialize $V^0,U^0$, $f$ and $\mathcal T_k$
                    \If{full-space correction}
                        \State \SthreeMath{P_s\gets0,\quad P_g\gets I,\quad\beta_s\gets0}
                        \State Freeze $\theta_g$ \Comment{Alg.~\ref{alg:score-learning}}
                    \Else
                        \State Fix $P_s$; $P_g\gets I-P_s$
                        \State Freeze $(\theta_s,\theta_g)$ \Comment{Alg.~\ref{alg:sequential-training}}
                    \EndIf
                    \Statex \Comment{Trained parameters remain fixed}
                    \For{$k=0,\ldots,K-1$}
                        \State \SthreeMath{A^{k+1}\gets\operatorname{prox}_{f/\rho}(V^k-U^k)}
                        \State \SthreeMath{C^k\gets A^{k+1}+U^k}
                        \State \SthreeMath{s_s^k\gets0,\quad s_g^k\gets0}
                        \For{$i\in\{s,g\}$ with $\beta_i>0$}
                            \State \SthreeMath{s_i^k\gets s_i(C^k,t_k;\theta_i)}
                        \EndFor
                        \State \SthreeMath{V^{k+1}\gets\mathcal T_k(C^k+\beta_s s_s^k+\beta_g s_g^k)}
                        \State \SthreeMath{U^{k+1}\gets U^k+A^{k+1}-V^{k+1}}
                    \EndFor
                    \State \Return $\widehat A=V^K$
                \end{algorithmic}
            \end{algorithm}
        \end{minipage}%
    }

    \setlength{\SthreeLeftGap}{\dimexpr
        \ht\SthreeInferenceBox+\dp\SthreeInferenceBox
        -\ht\SthreeLearnBox-\dp\SthreeLearnBox
        -\ht\SthreeSequentialBox-\dp\SthreeSequentialBox\relax}
    \ifdim\SthreeLeftGap<0pt
        \PackageWarning{SthreeAlgorithms}{Left algorithms exceed inference height; rebalance content}
        \setlength{\SthreeLeftGap}{0pt}
    \fi
    \typeout{STHREE-LEFT-GAP=\the\SthreeLeftGap}
    \noindent
    \vtop{\offinterlineskip\kern0pt
        \hbox{\usebox{\SthreeLearnBox}}%
        \vskip\SthreeLeftGap
        \hbox{\usebox{\SthreeSequentialBox}}%
    }\hfill
    \vtop{\offinterlineskip\kern0pt
        \hbox{\usebox{\SthreeInferenceBox}}%
    }
\end{figure}

\section{Experiments}
\label{sec:experiments}

We evaluate \methodname{} on reference-guided HSI restoration, single-image HSI denoising, and MRI denoising. The experiments include quantitative and qualitative comparisons, ablation studies, and controlled tests of the theoretical predictions.

\subsection{Datasets and Experimental Settings}
\label{sec:experimental-settings}

\noindent\textbf{Datasets and Experimental Settings.} We evaluate reference-guided hyperspectral restoration on the Dongying dataset from HiPandas \citep{xu2025hipandas}, using a noisy low-resolution HSI as input and a PAN image as guidance. For single-image hyperspectral denoising, we use Qingpu and Xiong'an, with a single noisy HSI as input. Both tasks are evaluated under Gaussian and mixed noise at varying intensity levels. MRI denoising is evaluated on BrainWeb at the official 3\%, 5\%, and 7\% noise levels and on M4Raw using acquired noisy images from all 25 test subjects. For all learning-based methods without pretrained weights, self-supervised training is performed at test time using only the noisy test observations. Dataset and implementation details are provided in Appendix~\ref{app:experimental-settings}.  We report PSNR and SSIM for all tasks, supplemented by SAM and ERGAS for both hyperspectral tasks; higher PSNR/SSIM and lower SAM/ERGAS indicate better performance.

\noindent\textbf{Baselines.} For reference-guided hyperspectral restoration, we follow the baseline setup of UHiPandas \citep{xu2025hipandas}. Comparisons include UHiPandas and cascaded pipelines constructed from R-PNN \citep{guarino2024rpnn}, HIR-Diff \citep{pang2024hirdiff}, PWRCTV \citep{xu2024pwrctv}, PLRDiff \citep{rui2024plrdiff}, and SwinIR \citep{liang2021swinir}. For single-image hyperspectral denoising, baselines include BM4D \citep{maggioni2013bm4d}, HLRTF \citep{luo2022hlrtf}, RCTV \citep{peng2022rctv}, DDS2M \citep{miao2023dds2m}, LRTFR \citep{luo2024lrtfr}, HIR-Diff \citep{pang2024hirdiff}, and CF-INR \citep{yu2026cfinr}. For MRI denoising, baselines include Noise2Self \citep{batson2019noise2self}, Noise2Void \citep{krull2019noise2void}, Noisier2Noise \citep{moran2020noisier2noise}, R2R \citep{pang2021r2r}, Noise2Score \citep{kim2021noise2score}, ZS-N2N \citep{mansour2023zsn2n}, C2S \citep{tu2025corruption2self}, and DRDD \citep{lin2026drdd}. In our comparisons, HIR-Diff, PLRDiff, and DRDD use pretrained diffusion weights, whereas DDS2M performs self-supervised learning during reverse diffusion and C2S trains a denoising predictor through generalized denoising score matching.

\subsection{Quantitative and Qualitative Results}
\label{sec:restoration-results}

\noindent\textbf{Quantitative comparisons.}
Tables~\ref{tab:hipandas-results}--\ref{tab:mri-denoising-results} and Appendix~\ref{app:additional-restoration-quantitative} summarize the results for the three restoration tasks. For reference-guided HSI restoration, \methodname{} achieves better average restoration performance than the compared cascaded methods and UHiPandas in both noise groups, with improved spectral fidelity under mixed noise. Similarly, for single-image HSI denoising on Qingpu, it outperforms the reported baselines under both Gaussian and mixed noise. For MRI denoising, \methodname{} achieves the best reported average performance at all three BrainWeb noise levels. On M4Raw, \methodname{} learns only at test time and achieves results comparable to pretrained C2S. \methodname{} generally outperforms the evaluated pretrained diffusion methods and the self-supervised diffusion method.

\begin{table}[!htbp]
\centering
\normalsize
\setlength{\tabcolsep}{2pt}
\renewcommand{\arraystretch}{1.02}
\setlength{\abovecaptionskip}{4pt}
\setlength{\belowcaptionskip}{0pt}
\caption{Reference-guided HSI restoration on Dongying. Best in bold; second best underlined.}
\label{tab:hipandas-results}
\resizebox{0.9\linewidth}{!}{%
\begin{tabular*}{\linewidth}{@{\extracolsep{\fill}}*{9}{c}@{}}
\toprule
& \multicolumn{4}{c}{\textbf{Gaussian Mean}}
& \multicolumn{4}{c}{\textbf{Mixed Mean}} \\
\cmidrule(lr){2-5}\cmidrule(lr){6-9}
Method & PSNR $\uparrow$ & SSIM $\uparrow$
& SAM $\downarrow$ & ERGAS $\downarrow$
& PSNR $\uparrow$ & SSIM $\uparrow$
& SAM $\downarrow$ & ERGAS $\downarrow$ \\
\midrule
R-PNN + HIR-Diff
& 30.83 & 0.7226 & 8.80 & 22.55
& 29.83 & 0.7139 & 11.44 & 31.40 \\
HIR-Diff + SwinIR
& 32.42 & 0.7431 & 7.00 & 19.42
& 29.76 & 0.6776 & 11.33 & 34.18 \\
HIR-Diff + R-PNN
& 35.32 & 0.9199 & 6.08 & 15.35
& 33.12 & 0.8995 & 9.75 & \underline{29.67} \\
PWRCTV + PLRDiff
& 36.13 & 0.8675 & 6.23 & 21.54
& 32.76 & 0.7747 & 13.59 & 153.69 \\
PWRCTV + SwinIR
& 35.55 & 0.8773 & 4.66 & 14.65
& 33.80 & 0.8403 & 9.90 & 160.43 \\
UHiPandas
& \underline{38.24} & \underline{0.9368}
& \underline{4.05} & \underline{10.87}
& \underline{37.17} & \underline{0.9294}
& \underline{7.10} & 32.91 \\
\midrule
\methodname{} (Ours)
& \textbf{39.75} & \textbf{0.9497}
& \textbf{3.45} & \textbf{9.20}
& \textbf{38.38} & \textbf{0.9438}
& \textbf{4.53} & \textbf{11.24} \\
\bottomrule
\end{tabular*}}
\end{table}

\begin{table}[!htbp]
\centering
\normalsize
\setlength{\tabcolsep}{2pt}
\renewcommand{\arraystretch}{1.02}
\setlength{\abovecaptionskip}{4pt}
\setlength{\belowcaptionskip}{0pt}
\caption{Single-image HSI denoising on Qingpu. Best in bold; second best underlined.}
\label{tab:hsi-seven-qingpu}
\resizebox{0.9\linewidth}{!}{%
\begin{tabular*}{\linewidth}{@{\extracolsep{\fill}}*{9}{c}@{}}
\toprule
& \multicolumn{4}{c}{\textbf{Gaussian Mean}}
& \multicolumn{4}{c}{\textbf{Mixed Mean}} \\
\cmidrule(lr){2-5}\cmidrule(lr){6-9}
Method & PSNR $\uparrow$ & SSIM $\uparrow$ & SAM $\downarrow$ & ERGAS $\downarrow$
& PSNR $\uparrow$ & SSIM $\uparrow$ & SAM $\downarrow$ & ERGAS $\downarrow$ \\
\midrule
BM4D & 33.95 & 0.8214 & 3.61 & 9.18 & 24.56 & 0.4833 & 17.50 & 68.67 \\
HLRTF & 34.94 & 0.7733 & 3.07 & 8.97 & 31.23 & 0.7944 & 27.69 & 117.96 \\
RCTV & 28.59 & 0.5180 & 7.34 & 19.50 & 34.01 & 0.8741 & 5.49 & 16.13 \\
DDS2M & 39.97 & 0.9504 & 1.40 & 4.21 & 29.79 & 0.8536 & 13.12 & 53.51 \\
LRTFR & 37.24 & 0.9020 & 1.78 & 6.17 & 30.84 & 0.8442 & 27.00 & 114.55 \\
HIR-Diff & 38.47 & 0.9157 & 1.68 & 4.85 & 29.06 & 0.8260 & 13.20 & 55.59 \\
CF-INR & \underline{40.32} & \underline{0.9541} & \underline{1.36} & \underline{4.01} & \underline{35.86} & \underline{0.9339} & \underline{2.62} & \underline{8.50} \\
\midrule
\methodname{} (Ours)
& \textbf{40.47} & \textbf{0.9562} & \textbf{1.33} & \textbf{3.95}
& \textbf{37.88} & \textbf{0.9433} & \textbf{1.89} & \textbf{5.34} \\
\bottomrule
\end{tabular*}%
}
\end{table}

\begin{table}[!htbp]
\centering
\normalsize
\setlength{\tabcolsep}{6.5pt}
\renewcommand{\arraystretch}{1.02}
\setlength{\abovecaptionskip}{4pt}
\setlength{\belowcaptionskip}{0pt}
\caption{MRI denoising. Best in bold; second best underlined.}
\label{tab:mri-denoising-results}
\resizebox{\linewidth}{!}{%
\begin{tabular}{@{}*{13}{c}@{}}
\toprule
 & \multicolumn{6}{c}{\textbf{BrainWeb}}
 & \multicolumn{6}{c}{\textbf{M4Raw}} \\
\cmidrule(lr){2-7}\cmidrule(lr){8-13}
Method
 & \multicolumn{2}{c}{3\%}
 & \multicolumn{2}{c}{5\%}
 & \multicolumn{2}{c}{7\%}
 & \multicolumn{2}{c}{T1}
 & \multicolumn{2}{c}{T2}
 & \multicolumn{2}{c}{FLAIR} \\
\cmidrule(lr){2-3}\cmidrule(lr){4-5}
\cmidrule(lr){6-7}\cmidrule(lr){8-9}
\cmidrule(lr){10-11}\cmidrule(lr){12-13}
 & PSNR & SSIM & PSNR & SSIM & PSNR & SSIM
 & PSNR & SSIM & PSNR & SSIM & PSNR & SSIM \\
\midrule
Noise2Self
 & 31.22 & 0.8613 & 30.40 & 0.8396 & 29.49 & 0.7952
 & 29.10 & 0.8537 & 28.17 & 0.8095 & 28.94 & 0.8150 \\
Noise2Void
 & 32.11 & 0.8810 & 30.40 & \underline{0.8522} & 29.23 & 0.8110
 & 31.24 & 0.8671 & 30.45 & 0.8392 & 30.80 & 0.8288 \\
Noisier2Noise
 & 34.38 & 0.8964 & 31.37 & 0.8341 & 28.90 & 0.7410
 & 31.18 & 0.8675 & 30.55 & 0.8435 & 31.17 & 0.8356 \\
R2R
 & 34.02 & 0.8011 & 32.03 & 0.7879 & 29.50 & 0.7292
 & 30.82 & 0.8552 & 30.48 & 0.8329 & 30.97 & 0.8187 \\
Noise2Score
 & 34.47 & 0.8405 & 29.95 & 0.7621 & 26.85 & 0.7133
 & 31.23 & 0.8444 & 30.48 & 0.8139 & 31.00 & 0.7975 \\
ZS-N2N
 & 32.63 & 0.7768 & 28.73 & 0.7165 & 26.07 & 0.6749
 & 31.66 & 0.8922 & 30.96 & 0.8729 & 31.89 & 0.8765 \\
C2S
 & \underline{35.21} & \underline{0.9287}
 & \underline{33.30} & 0.8504
 & \underline{31.88} & \underline{0.8173}
 & \underline{32.61} & \underline{0.9152}
 & \underline{32.30} & \underline{0.8880}
 & \textbf{33.02} & 0.8792 \\
DRDD
 & 35.03 & 0.8188 & 30.70 & 0.7581 & 27.83 & 0.7233
 & 32.43 & 0.9108 & 31.44 & 0.8770 & 32.60 & \underline{0.8797} \\
\midrule
\methodname{} (Ours)
 & \textbf{37.47} & \textbf{0.9568}
 & \textbf{34.65} & \textbf{0.9413}
 & \textbf{32.72} & \textbf{0.9279}
 & \textbf{32.69} & \textbf{0.9165}
 & \textbf{32.31} & \textbf{0.8922}
 & \underline{32.83} & \textbf{0.8906} \\
\bottomrule
\end{tabular}%
}
\end{table}

\begin{figure}[!htbp]
    \centering
    \resizebox{\linewidth}{!}{%
        \includegraphics{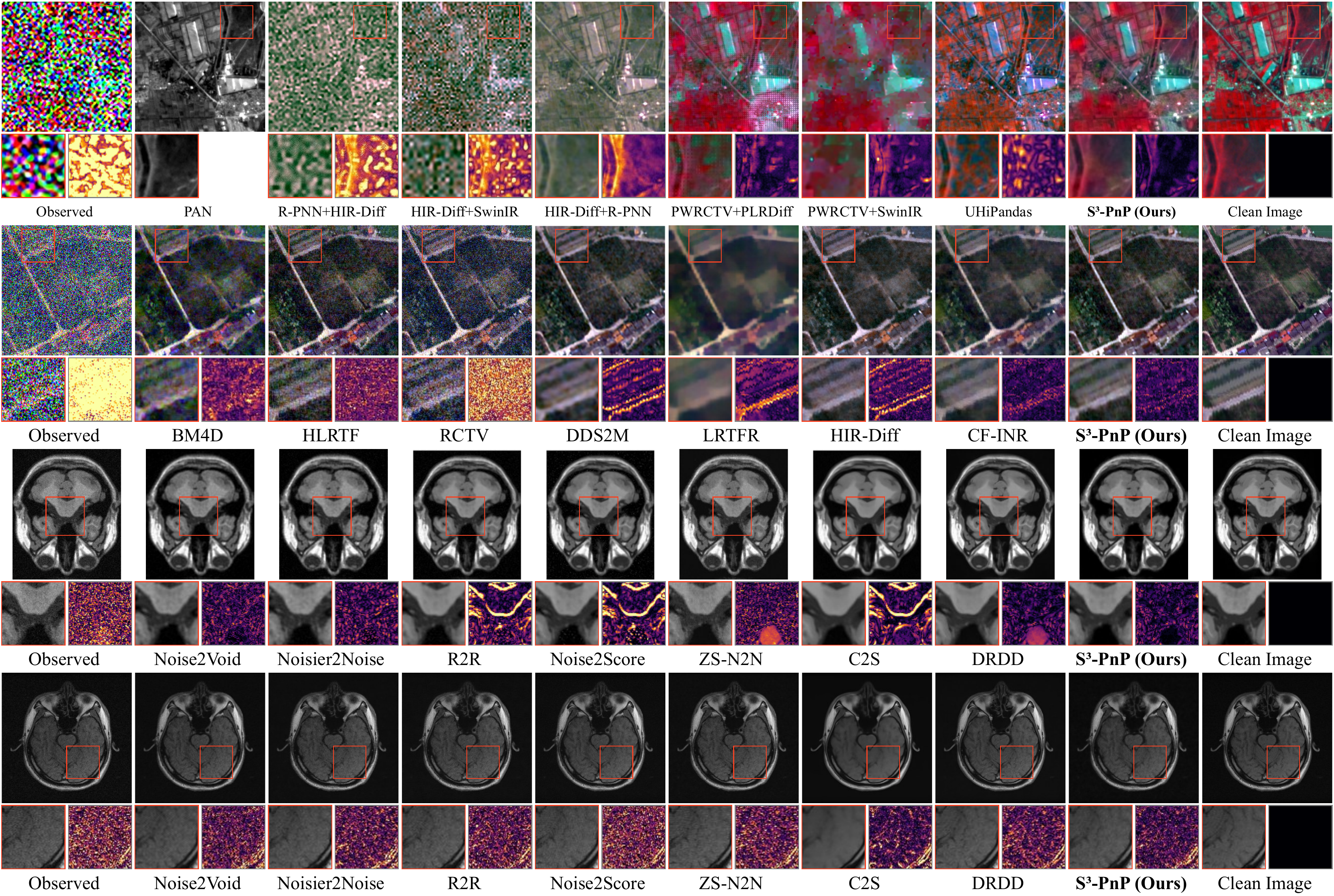}%
    }
    \caption{Qualitative comparisons on Dongying (GNI), Qingpu (G50), BrainWeb (T1, 7\% noise), and M4Raw (T1), from top to bottom. Enlarged regions and corresponding error maps are shown below each reconstruction.}
    \label{fig:qualitative-results}
\end{figure}

\noindent\textbf{Qualitative comparisons.}
Figure~\ref{fig:qualitative-results} presents visual comparisons for the three restoration tasks. For reference-guided HSI restoration on Dongying, \methodname{} recovers clearer spatial boundaries and colors closer to the reference, while several cascaded methods exhibit color shifts or residual artifacts. For single-image HSI denoising on Qingpu, it suppresses noise while preserving fine spatial structures, whereas some competing methods leave visible noise or blur local details. On BrainWeb and M4Raw, it reduces noise while retaining tissue boundaries and local anatomical structures. The enlarged regions highlight these differences in noise suppression and detail preservation, and the error maps reveal the remaining discrepancies from the references.

\subsection{Ablation Studies}
\label{sec:ablations}

\begin{figure}[!htbp]
    \centering
    \includegraphics[width=\linewidth]{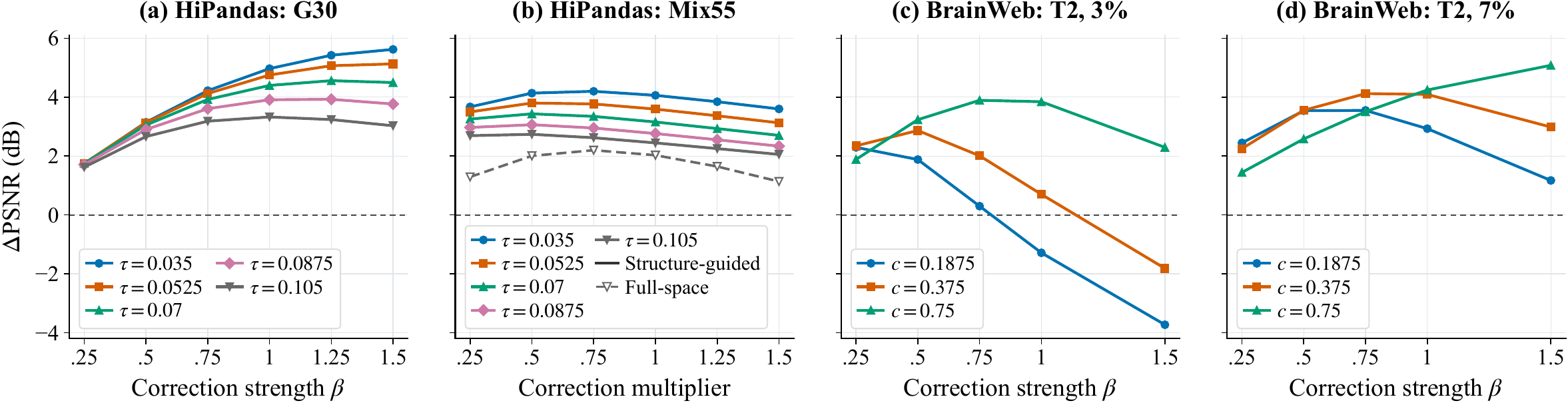}
    \caption{PSNR gains over analytical ADMM at the same prior setting, varying $\tau$ in (a)--(b) and $c$ in (c)--(d). In (b), solid curves denote structure-guided correction and the dashed curve denotes full-space correction at $\tau=0.105$; the horizontal axis scales their respective reference gains. BrainWeb results average three development slices per condition.}
    \label{fig:ablation-correction-strength}
\end{figure}

\noindent\textbf{Contribution of learned corrections.}
We compare each trained configuration with analytical ADMM at the same prior setting. The analytical reference uses the initial observation-derived estimates, and network weights remain fixed during each correction-strength sweep. Figure~\ref{fig:ablation-correction-strength}(a)--(b) shows positive gains throughout the tested Dongying grids. On BrainWeb, all tested settings improve performance at 7\% noise, while strong corrections under weaker priors can reduce performance at 3\% noise (c)--(d). These results show that learned corrections can improve analytical ADMM, but their strength should be selected jointly with the analytical prior and noise level.

\noindent\textbf{Structure-guided learning.}
We compare structure-guided and full-space learning at the same prior strength $\tau=0.105$. The structure-guided configuration achieves larger gains across all tested correction multipliers (Figure~\ref{fig:ablation-correction-strength}(b)). Since the training configurations also differ, this comparison does not isolate the contribution of orthogonal projection alone.

\subsection{Consistency with the Theoretical Analysis}
\label{sec:theory-experiments}

\begin{figure}[!htbp]
    \centering
    \includegraphics[width=\linewidth]{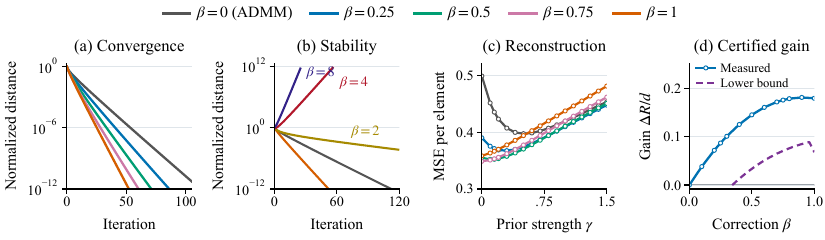}
    \caption{Controlled experiments with $64\times64$ Gaussian states and ideal scores. (a)--(c) Wide-spectrum fidelity; (d) near-matched fidelity with a theoretical lower bound on error reduction. The relative prior strength $\gamma=b/a_t$ is fixed at $0.15$ in (a)--(b) and $0.1$ in (d). Shading in (c)--(d) indicates pointwise 95\% confidence intervals.}
    \label{fig:theory-beta}
\end{figure}

We examine Theorem~\ref{prop:gdsm-recovery} using matched solvers with positive-definite quadratic fidelity and exact singular-value soft thresholding. Under wide-spectrum fidelity, the tested corrections accelerate the empirical decay of the normalized distance between trajectories from two initializations (Figure~\ref{fig:theory-beta}(a)): at $\beta=1$, the distance reaches $10^{-10}$ in 44 iterations, compared with 93 for analytical ADMM. Some settings outside the sufficient contraction region still show decaying distances, whereas $\beta=4,8$, beyond the theorem's range, produce growing distances (b).
Moderate corrections ($\beta=0.25,0.5$) reduce reconstruction error across the tested prior strengths, whereas stronger corrections can increase error under strong analytical regularization (c). Under near-matched fidelity, all eight sampled settings satisfying the improvement condition yield empirical gains above the positive theoretical lower bound (d), consistent with the predicted improvement and illustrating the bound's conservatism. These results support balancing score correction with analytical regularization. Appendix~\ref{app:theory-experiment-details} provides full settings and results across dimensions.

\section{Conclusion}
We presented \methodname{}, a self-supervised PnP-ADMM framework for scientific image restoration. By learning score corrections from ADMM states generated by the available observations, the framework connects prior learning to the restoration process. These corrections refine the inputs to analytical prior updates, complementing analytical regularization without clean training targets or externally pretrained weights. We further analyzed the convergence and reconstruction error of the corrected ADMM algorithm under ideal scores and quadratic data fidelity. Through task-specific observation models, the framework supports reference-guided HSI restoration, single-image HSI denoising, and MRI denoising. Our work provides a common approach to integrating observation-driven score learning with analytical optimization for scientific imaging.

\FloatBarrier
\begingroup
\interlinepenalty=10000

\subsection*{AI use statement}
In this work, we used generative AI tools solely to improve the grammar, clarity,
and readability of text written by the authors. Generative AI was not used to
generate research ideas, formulate hypotheses or claims, design the methodology
or experiments, implement the method, analyze data, or interpret results. The
authors reviewed all AI-assisted edits and take full responsibility for the final
content of this work.

\bibliography{references}
\bibliographystyle{iclr2027_conference}
\endgroup

\clearpage
\appendix
\numberwithin{equation}{section}

\section{Proofs}
\label{app:proofs}

\newcommand{\sThreeRestatedName}{}
\theoremstyle{plain}
\newtheorem*{sThreeRestated}{\sThreeRestatedName}
\theoremstyle{definition}

\subsection{Proof of Lemma~\ref{lem:state-gdsm}}
\label{app:proof-state-score}

\renewcommand{\sThreeRestatedName}{Lemma~\ref*{lem:state-gdsm}}
\begin{sThreeRestated}[GDSM regression on ADMM states; restated]
Let $C$ be an ADMM denoiser input drawn from a fixed reference training distribution, and let $Z=\mathcal S(C)=(C-m)/ s$, where the channel-wise median $m$ and positive scale $s$ are estimated from the training set and fixed; $/$ denotes element-wise division. Assume $\mathbb E\|Z\|^2<\infty$. For fixed $t$, construct $Z_t=Z+\sqrt{\Delta_t}\epsilon$, where $\Delta_t>0$ and $\epsilon\sim\mathcal N(0,I)$ is independent of $Z$. Let $p_t$ denote the density of $Z_t$ and define $R_t=-\log p_t$. For $\kappa_t>\Delta_t$, consider the GDSM interpolation loss \citep{tu2025corruption2self}
\[
    \mathcal L_{\mathrm{GDSM}}(\widehat Z;t)
    =
    \mathbb E\!\left\|
        \frac{\Delta_t}{\kappa_t}\widehat Z(Z_t,t)
        +\left(1-\frac{\Delta_t}{\kappa_t}\right)Z_t-Z
    \right\|^2.
\]
Its minimizer over square-integrable restoration predictors satisfies, almost surely,
\[
    \widehat Z^\star(Z_t,t)
    =
    Z_t+\kappa_t s_t^\star(Z_t),
\]
where $s_t^\star=\nabla\log p_t=-\nabla R_t$. Moreover, $\mathbb E[Z\mid Z_t]=Z_t+\Delta_t s_t^\star(Z_t)$. If additionally $Z=Z_{\mathrm{0}}+\sigma_0\epsilon_0$, where $\sigma_0>0$, $\epsilon_0\sim\mathcal N(0,I)$, and $Z_{\mathrm{0}},\epsilon_0,\epsilon$ are mutually independent, choosing $\kappa_t=\sigma_0^2+\Delta_t$ gives $\widehat Z^\star(Z_t,t)=\mathbb E[Z_{\mathrm{0}}\mid Z_t]$. For conditional networks, all densities, expectations, and independence assumptions are understood conditional on the same auxiliary inputs.
\end{sThreeRestated}

\begin{proof}
Fix $t$, the reference distribution, and the normalization. Any auxiliary inputs are held fixed throughout. Assume $\mathbb E\|Z\|^2<\infty$, and let $Z_t=Z+\sqrt{\Delta_t}\epsilon$, where $\epsilon\sim\mathcal N(0,I)$ is independent of $Z$. We first characterize the optimal interpolated prediction, then derive its score representation using Tweedie's identity, and finally establish the clean-state conditional-mean interpretation.

\medskip
\noindent\textbf{Determine the optimal interpolated prediction.}
Consider the GDSM loss over all square-integrable restoration predictors $\widehat Z$:
\[
    \mathcal L_{\mathrm{GDSM}}(\widehat Z;t)
    =
    \mathbb E\!\left\|
        \frac{\Delta_t}{\kappa_t}\widehat Z(Z_t,t)
        +\left(1-\frac{\Delta_t}{\kappa_t}\right)Z_t-Z
    \right\|^2.
\]
Since $\Delta_t/\kappa_t>0$ and $Z_t$ is square-integrable, the affine transformation from the restoration prediction to its interpolation is invertible within the class of square-integrable predictors. Minimizing the GDSM loss is therefore equivalent to regressing the interpolated prediction onto $Z$ under squared error. By the conditional-mean characterization of optimal squared-error regression, the minimizer satisfies
\begin{equation}
    \mathbb E[Z\mid Z_t]
    =
    \frac{\Delta_t}{\kappa_t}\widehat Z^\star(Z_t,t)
    +\left(1-\frac{\Delta_t}{\kappa_t}\right)Z_t.
    \label{eq:app-optimal-interpolation}
\end{equation}
This determines the optimal interpolation. We next solve for the restoration prediction itself.

\medskip
\noindent\textbf{Derive the score representation of the optimal prediction.}
Let $p_t$ denote the density of $Z_t$ and define $s_t^\star=\nabla\log p_t$. For the independent Gaussian perturbation above, Tweedie's identity \citep{efron2011tweedie} gives
\begin{equation}
    \mathbb E[Z\mid Z_t]
    =
    Z_t+\Delta_t s_t^\star(Z_t).
    \label{eq:app-state-tweedie}
\end{equation}
Substituting this identity into Eq.~\eqref{eq:app-optimal-interpolation} and isolating $\widehat Z^\star$ yields
\begin{equation}
    \begin{aligned}
    \widehat Z^\star(Z_t,t)
    &=
    \frac{\kappa_t}{\Delta_t}\mathbb E[Z\mid Z_t]
    -\left(\frac{\kappa_t}{\Delta_t}-1\right)Z_t\\
    &=
    Z_t+\kappa_t s_t^\star(Z_t).
    \end{aligned}
    \label{eq:app-optimal-gdsm-prediction}
\end{equation}
Thus, the optimal restoration predictor takes the form of the input plus a scaled state score. This form follows from minimizing the interpolation loss, without imposing a score parameterization beforehand. Furthermore, defining $R_t=-\log p_t$ gives $s_t^\star=-\nabla R_t$, establishing the state-energy interpretation of the ideal correction.

\medskip
\noindent\textbf{Establish the clean-state conditional-mean interpretation.}
Under the additional assumption $Z=Z_{\mathrm{0}}+\sigma_0\epsilon_0$, where $Z_{\mathrm{0}}$, $\epsilon_0$, and $\epsilon$ are mutually independent and $\epsilon_0\sim\mathcal N(0,I)$, we have
\[
    Z_t
    =
    Z_{\mathrm{0}}
    +\sigma_0\epsilon_0
    +\sqrt{\Delta_t}\epsilon.
\]
The clean state is square-integrable because $Z_{\mathrm{0}}=Z-\sigma_0\epsilon_0$. The combined noise is Gaussian, independent of $Z_{\mathrm{0}}$, and has covariance $(\sigma_0^2+\Delta_t)I$. Applying Tweedie's identity to this observation model therefore gives
\[
    \mathbb E[Z_{\mathrm{0}}\mid Z_t]
    =
    Z_t+(\sigma_0^2+\Delta_t)s_t^\star(Z_t).
\]
The score is unchanged because the marginal distribution of $Z_t$ remains $p_t$. Choosing $\kappa_t=\sigma_0^2+\Delta_t$ and using Eq.~\eqref{eq:app-optimal-gdsm-prediction}, we conclude that
\[
    \widehat Z^\star(Z_t,t)
    =
    \mathbb E[Z_{\mathrm{0}}\mid Z_t].
\]
This proves the final statement. Neither the clean-state distribution nor $p_t$ is required to be Gaussian. Thus, the optimal GDSM prediction equals the input plus a scaled state score and, under the additional Gaussian residual assumption and matched coefficient, coincides with the conditional mean of the clean state.
\end{proof}

\subsection{Proof of Theorem~\ref{prop:gdsm-recovery}}
\label{app:proof-gdsm-recovery}

We restate the statistical construction and give the constants used in the theorem. Let $Z=Z_0+\sigma_0\epsilon_0$ and $Z_t=Z+\sqrt{\Delta_t}\epsilon$, where $\sigma_0,\Delta_t>0$, the two noises are standard Gaussian, and $Z_0,\epsilon_0,\epsilon$ are mutually independent. Assume $\mathbb E\|Z\|^2<\infty$. Fix the reference distribution, smoothing level, normalization $\mathcal S(C)=(C-m)/ s$ with $s>0$, and analytical map throughout the iterations and statistical expectations. All statements are conditional on any fixed auxiliary inputs.

The clean state in the original coordinates is $A_0=\mathcal S^{-1}(Z_0)$, and its noisy input is $Y=\mathcal S^{-1}(Z_t)$. Thus, the subscript $0$ in $A_0$ and $Z_0$ denotes the same clean state in different coordinates. Define the posterior-mean estimator $\bar A(Y)=\mathbb E[A_0\mid Y]$, its distance from the input $a_t=(\mathbb E\|Y-\bar A(Y)\|^2)^{1/2}$, and the matched coefficient $\kappa_t=\sigma_0^2+\Delta_t$. The ideal score $s_t^\star=\nabla\log p_t$ induces the original-coordinate correction and prior map
\[
    d_t(C)=\kappa_t s\odot s_t^\star(\mathcal S(C)),
    \qquad
    \mathcal C_\beta(C)=\mathcal T(C+\beta d_t(C)),
    \qquad \beta\in[0,1].
\]
This is Eq.~\eqref{eq:g-interface} with $T_k \equiv T$, $t_k \equiv t$,
$\beta_k \equiv \beta$, and $s_\theta(\cdot,t)=s_t^\star$;
$\beta=0$ disables the correction. Matrices are treated as elements of a finite-dimensional Hilbert space with the Frobenius inner product, and $\|\cdot\|_{\mathrm{op}}$ denotes the induced operator norm.

\renewcommand{\sThreeRestatedName}{Theorem~\ref*{prop:gdsm-recovery}}
\begin{sThreeRestated}[Convergence and error improvement under quadratic fidelity]
Under the construction above, use exact ADMM with fixed penalty $\rho>0$ and $f(A)=\frac12\langle A-Y,H(A-Y)\rangle$, where $H=H^\ast\succ0$ is fixed. Suppose $d_t$ is $L$-Lipschitz and $\mathcal T$ is firmly nonexpansive. Define $r_H=\|(\rho I-H)(\rho I+H)^{-1}\|_{\mathrm{op}}$. For any $\beta\in[0,1]$ satisfying
\begin{equation}
    r_H(1+2\beta L)<1,
    \label{eq:app-spd-stability}
\end{equation}
the ADMM iterates converge from any finite initialization, with
\begin{equation}
    \|C_\beta^k-C_\beta^\star\|
    \le q_\beta^k\|C_\beta^0-C_\beta^\star\|,
    \qquad q_\beta=\tfrac12[1+r_H(1+2\beta L)]<1.
    \label{eq:gdsm-admm-limit}
\end{equation}
Write $A_\beta^\star=\lim_k A_\beta^k=\lim_k V_\beta^k$ for the primal limit. Denote the limit with $\beta=0$ by $A_{\mathrm{off}}^\star$, to distinguish the correction-disabled estimate from the clean state $A_0$.

For the risk comparison, additionally assume $\|\mathcal T(Q)-Q\|\le b$ for all $Q$. Set $M=\rho H^{-1}$, $b_H=\|M\|_{\mathrm{op}}b$, and $\delta_\beta=\beta L\|M\|_{\mathrm{op}}\|I-H/\rho\|_{\mathrm{op}}$. When $\delta_\beta<1$, define $c_\beta=\|I-\beta M\|_{\mathrm{op}}+\beta\|M\|_{\mathrm{op}}\delta_\beta/(1-\delta_\beta)$. For $\beta>0$, the sufficient gain conditions
\begin{equation}
    \delta_\beta<1,
    \qquad
    c_\beta a_t+\frac{b_H}{1-\delta_\beta}<a_t-b_H
    \label{eq:app-spd-gain-condition}
\end{equation}
ensure
\[
    \mathbb E\|A_\beta^\star-A_0\|^2
    <\mathbb E\|A_{\mathrm{off}}^\star-A_0\|^2.
\]
The two solvers use the same data term, analytical map, and penalty; only the score correction changes.
\end{sThreeRestated}

\begin{proof}
The proof has three parts. First, we identify the useful correction supplied by the ideal score. Second, we establish convergence of the complete ADMM iteration. Third, we control the effect of the analytical map and ADMM feedback to compare the limiting reconstruction risks.

\par\smallskip\noindent\textbf{Relating the ideal score correction to the posterior mean.}
Lemma~\ref{lem:state-gdsm} gives $\mathbb E[Z_0\mid Z_t]=Z_t+\kappa_t s_t^\star(Z_t)$ under the matched coefficient. Since $\mathcal S$ is fixed, affine, and invertible, conditioning on $Y$ is equivalent to conditioning on $Z_t$.

Here $d_t(Y)=\kappa_t s\odot s_t^\star(\mathcal S(Y))$ denotes the ideal score correction rescaled to the original coordinates, with $s$ denoting the fixed normalization scale. Inverse normalization multiplies by $s$ and adds $m$, so the normalized input-plus-score prediction becomes $Y+d_t(Y)$. Using this rescaling and the fact that a fixed affine map commutes with conditional expectation, we obtain
\begin{equation}
    \begin{aligned}
        \bar A(Y)
        &=\mathbb E[A_0\mid Y]\\
        &=\mathcal S^{-1}\!\left[\mathbb E[Z_0\mid Z_t]\right]\\
        &=\mathcal S^{-1}\!\left[
          \mathcal S(Y)+\kappa_t s_t^\star(\mathcal S(Y))\right]\\
        &=Y+\kappa_t s\odot s_t^\star(\mathcal S(Y))
          =Y+d_t(Y).
    \end{aligned}
    \label{eq:app-physical-posterior}
\end{equation}
Thus, $d_t(Y)=\bar A(Y)-Y$: the ideal correction at $Y$ points toward the posterior mean, which minimizes expected squared reconstruction error. Substituting this identity gives
\begin{equation}
    Y+\beta d_t(Y)=(1-\beta)Y+\beta\bar A(Y)=:M_\beta.
    \label{eq:app-posterior-interpolation}
\end{equation}
The moment assumption makes $A_0$, $Y$, and $\bar A(Y)$ square-integrable, and $\mathbb E\|d_t(Y)\|^2=a_t^2$. The quadratic fidelity is centered at this same $Y$, so the correction identified here is calibrated to the input used in the risk comparison.

\par\smallskip\noindent\textbf{Establishing ADMM convergence.}
Fix $\beta\in[0,1]$ satisfying Eq.~\eqref{eq:app-spd-stability}. Suppress the subscript $\beta$ on intermediate iterates for readability. The prior update is
\begin{equation}
    V^{k+1}=\mathcal C_\beta(C^k)
    =\mathcal T(C^k+\beta d_t(C^k)).
    \label{eq:app-general-prior-update}
\end{equation}
Write $h=HY$. The first-order condition for the exact data update is $HA-h+\rho(A-V^k+U^k)=0$, giving
\begin{equation}
    A^{k+1}=(H+\rho I)^{-1}\bigl[h+\rho(V^k-U^k)\bigr].
    \label{eq:app-spd-data-update}
\end{equation}
Using $C^k=A^{k+1}+U^k$ and $U^{k+1}=C^k-V^{k+1}$, the next prior input satisfies
\begin{equation}
    \begin{aligned}
        C^{k+1}
        &=A^{k+2}+U^{k+1}\\
        &=(H+\rho I)^{-1}
          \bigl[h+\rho V^{k+1}+HU^{k+1}\bigr]\\
        &=(H+\rho I)^{-1}
          \bigl[h+HC^k+(\rho I-H)\mathcal C_\beta(C^k)\bigr]
          =:\mathcal F_\beta(C^k).
    \end{aligned}
    \label{eq:app-input-recurrence}
\end{equation}
To prove convergence, we show that the complete ADMM map $\mathcal F_\beta$ contracts distances between inputs. We use a reflected-operator representation \citep{ryu2019pnpconvergence} to exploit the firm nonexpansiveness of $\mathcal T$: its reflection $2\mathcal T-I$ is nonexpansive, which allows us to bound the effect of the added score correction. Define $R(v)=2(H+\rho I)^{-1}(h+\rho v)-v$ and $R_\beta=2\mathcal C_\beta-I$ as the reflections of the data and prior updates, respectively. Substitution gives
\[
    \begin{aligned}
        \tfrac12\bigl[C+R(R_\beta(C))\bigr]
        &=(H+\rho I)^{-1}\bigl[h+\rho R_\beta(C)\bigr]
          +\tfrac12\bigl[C-R_\beta(C)\bigr]\\
        &=(H+\rho I)^{-1}\bigl[h+\rho(2\mathcal C_\beta(C)-C)\bigr]
          +C-\mathcal C_\beta(C)\\
        &=(H+\rho I)^{-1}\bigl[h+HC+(\rho I-H)\mathcal C_\beta(C)\bigr]
          =\mathcal F_\beta(C).
    \end{aligned}
\]
Thus Eq.~\eqref{eq:app-input-recurrence} has the equivalent representation
\begin{equation}
    \begin{aligned}
        \mathcal F_\beta&=\tfrac12(I+R\circ R_\beta),\\
        R(v)&=(\rho I-H)(\rho I+H)^{-1}v
                +2(H+\rho I)^{-1}h.
    \end{aligned}
    \label{eq:app-spd-reflection}
\end{equation}
We now bound the changes in $R$ and $R_\beta$ separately to establish contraction of $\mathcal F_\beta$. Positive definiteness of $H$ implies
\[
    \operatorname{Lip}(R)=r_H
    =\max_j\left|\frac{\rho-\lambda_j(H)}{\rho+\lambda_j(H)}\right|<1.
\]
Firm nonexpansiveness of $\mathcal T$ means $\|\mathcal T(x)-\mathcal T(y)\|^2\le\langle\mathcal T(x)-\mathcal T(y),x-y\rangle$, and therefore $R_{\mathcal T}=2\mathcal T-I$ is nonexpansive. Using $\mathcal C_\beta(C)=\mathcal T(C+\beta d_t(C))$ and adding and subtracting the corrected input yields
\begin{equation}
    \begin{aligned}
        R_\beta(C)
        &=2\mathcal C_\beta(C)-C\\
        &=2\mathcal T(C+\beta d_t(C))-C\\
        &=\bigl[2\mathcal T(C+\beta d_t(C))-(C+\beta d_t(C))\bigr]
          +\beta d_t(C)\\
        &=R_{\mathcal T}(C+\beta d_t(C))+\beta d_t(C).
    \end{aligned}
    \label{eq:app-spd-prior-reflection}
\end{equation}
For any two inputs $C,C'$, the triangle inequality and nonexpansiveness of $R_{\mathcal T}$ then give
\begin{equation}
    \begin{aligned}
        &\|R_\beta(C)-R_\beta(C')\|\\
        &\le\|R_{\mathcal T}(C+\beta d_t(C))-R_{\mathcal T}(C'+\beta d_t(C'))\|
            +\beta\|d_t(C)-d_t(C')\|\\
        &\le\|(C-C')+\beta(d_t(C)-d_t(C'))\|
            +\beta\|d_t(C)-d_t(C')\|\\
        &\le\|C-C'\|+2\beta\|d_t(C)-d_t(C')\|\\
        &\le(1+2\beta L)\|C-C'\|.
    \end{aligned}
    \label{eq:app-spd-prior-lipschitz}
\end{equation}
The last inequality uses the $L$-Lipschitz property of $d_t$. Thus $\operatorname{Lip}(R_\beta)\le1+2\beta L$. Combining this bound with $\operatorname{Lip}(R)=r_H$ in Eq.~\eqref{eq:app-spd-reflection}, we obtain
\begin{equation}
    \begin{aligned}
        &\|\mathcal F_\beta(C)-\mathcal F_\beta(C')\|\\
        &=\tfrac12\bigl\|(C-C')+R(R_\beta(C))-R(R_\beta(C'))\bigr\|\\
        &\le\tfrac12\|C-C'\|
            +\tfrac12\|R(R_\beta(C))-R(R_\beta(C'))\|\\
        &\le\tfrac12\|C-C'\|
            +\tfrac{r_H}{2}\|R_\beta(C)-R_\beta(C')\|\\
        &\le\tfrac12\|C-C'\|
            +\tfrac{r_H}{2}(1+2\beta L)\|C-C'\|\\
        &=\tfrac12[1+r_H(1+2\beta L)]\|C-C'\|
          =q_\beta\|C-C'\|.
    \end{aligned}
    \label{eq:app-spd-contraction}
\end{equation}
The inequalities successively use the triangle inequality, the bound for $R$, and Eq.~\eqref{eq:app-spd-prior-lipschitz}. Since $q_\beta<1$, the contraction principle gives a unique fixed point $C_\beta^\star$ and the linear convergence bound in Eq.~\eqref{eq:gdsm-admm-limit}. Continuity of $\mathcal C_\beta$ and the ADMM updates then imply $A^k,V^k\to A_\beta^\star$ and $U^k\to C_\beta^\star-A_\beta^\star$, with
\begin{equation}
    A_\beta^\star=\mathcal C_\beta(C_\beta^\star),
    \qquad
    \rho C_\beta^\star=h+(\rho I-H)A_\beta^\star.
    \label{eq:app-spd-fixed-point}
\end{equation}
This also covers $\beta=0$, for which $q_0=(1+r_H)/2<1$.

\par\smallskip\noindent\textbf{Comparing the limiting reconstruction risks.}
Part 2 establishes the limiting estimates. We now compare their expected reconstruction errors using the posterior mean $\bar A(Y)=\mathbb E[A_0\mid Y]$ as a common reference. For any square-integrable estimator $\widehat A(Y)$, expanding $\widehat A(Y)-A_0=(\widehat A(Y)-\bar A(Y))+(\bar A(Y)-A_0)$ gives
\begin{equation}
    \begin{aligned}
        \mathbb E\|\widehat A(Y)-A_0\|^2
        &=R_{\mathrm{MMSE}}
          +\mathbb E\|\widehat A(Y)-\bar A(Y)\|^2,\\
        R_{\mathrm{MMSE}}&=\mathbb E\|\bar A(Y)-A_0\|^2.
    \end{aligned}
    \label{eq:app-bayes-risk}
\end{equation}
The cross term vanishes since $\mathbb E[\bar A(Y)-A_0\mid Y]=0$. The two ADMM limits are measurable functions of $Y$, as they can be obtained by iterating the fixed continuous maps from a fixed initialization. Their square integrability will be verified below. With this property, applying Eq.~\eqref{eq:app-bayes-risk} to both estimates cancels the common $R_{\mathrm{MMSE}}$ and gives
\begin{equation}
    \begin{aligned}
        &\mathbb E\|A_{\mathrm{off}}^\star-A_0\|^2
          -\mathbb E\|A_\beta^\star-A_0\|^2\\
        &\qquad=\mathbb E\|A_{\mathrm{off}}^\star-\bar A(Y)\|^2
          -\mathbb E\|A_\beta^\star-\bar A(Y)\|^2.
    \end{aligned}
    \label{eq:app-risk-comparison-target}
\end{equation}
Our goal is therefore to make the difference on the right strictly positive. We derive a lower bound for its first expectation and an upper bound for its second, then compare them using the gain condition. For the intermediate estimates, write $\|X\|_{L^2}:=(\mathbb E\|X\|^2)^{1/2}$, so $a_t=\|Y-\bar A(Y)\|_{L^2}$. We first bound these root-mean-square distances using triangle inequalities, then square the bounds to compare the expectations.

We begin with the correction-disabled estimate. To determine how far it can move from $Y$, we express the output displacement using the fixed-point equations from Part 2. Let $B=I-H/\rho$. Since $h=HY$, Eq.~\eqref{eq:app-spd-fixed-point} gives, for either solver,
\begin{equation}
    C_\beta^\star-Y=B(A_\beta^\star-Y).
    \label{eq:app-spd-state-shift}
\end{equation}
Define the displacement of the analytical map at its actual input by
\[
    e_\beta=\mathcal T(C_\beta^\star+\beta d_t(C_\beta^\star))
             -(C_\beta^\star+\beta d_t(C_\beta^\star)),
    \qquad \|e_\beta\|\le b.
\]
The prior fixed-point equation gives $A_\beta^\star-C_\beta^\star=\beta d_t(C_\beta^\star)+e_\beta$, while Eq.~\eqref{eq:app-spd-state-shift} gives $A_\beta^\star-C_\beta^\star=(H/\rho)(A_\beta^\star-Y)$. Combining the two identities yields
\begin{equation}
    A_\beta^\star-Y
    =M\bigl[\beta d_t(C_\beta^\star)+e_\beta\bigr],
    \qquad M=\rho H^{-1}.
    \label{eq:app-spd-output-shift}
\end{equation}
Setting $\beta=0$ gives $A_{\mathrm{off}}^\star-Y=Me_{\mathrm{off}}$, where $e_{\mathrm{off}}$ is the corresponding analytical-map displacement. Hence $\|A_{\mathrm{off}}^\star-Y\|\le b_H$, which also makes $A_{\mathrm{off}}^\star$ square-integrable because $Y$ is square-integrable. The reverse-triangle inequality now gives
\begin{equation}
    \begin{aligned}
        \|A_{\mathrm{off}}^\star-\bar A(Y)\|_{L^2}
        &\ge \|Y-\bar A(Y)\|_{L^2}
          -\|A_{\mathrm{off}}^\star-Y\|_{L^2}\\
        &\ge a_t-b_H.
    \end{aligned}
    \label{eq:app-off-posterior-distance}
\end{equation}
The gain condition ensures $a_t>b_H$, so squaring this nonnegative lower bound gives $\mathbb E\|A_{\mathrm{off}}^\star-\bar A(Y)\|^2\ge(a_t-b_H)^2$. This is the required lower bound for the first expectation in Eq.~\eqref{eq:app-risk-comparison-target}.

Nextly, we seek an upper bound for the corrected estimate's distance to the posterior mean. Part 1 identifies $d_t(Y)=\bar A(Y)-Y$, but at convergence ADMM applies the score correction to $C_\beta^\star$, which generally differs from $Y$. Adding and subtracting $\beta M d_t(Y)$ in Eq.~\eqref{eq:app-spd-output-shift} separates the calibrated score contribution from the remaining effects:
\[
    \begin{aligned}
        A_\beta^\star-Y&=\beta M d_t(Y)+\xi_\beta,\\
        \xi_\beta&=\beta M\bigl[d_t(C_\beta^\star)-d_t(Y)\bigr]+Me_\beta.
    \end{aligned}
\]
Substituting the identity from Part 1 gives
\[
    A_\beta^\star-\bar A(Y)
    =(I-\beta M)(Y-\bar A(Y))+\xi_\beta.
\]
The first term describes how the calibrated score contribution changes the original discrepancy. The remainder $\xi_\beta$ contains the change in score correction caused by the shifted ADMM input and the displacement due to the analytical map. Bounding this remainder will give the desired upper bound.

For $\beta>0$, set $D_\beta=A_\beta^\star-Y$. To control the remainder, we first bound this output displacement by separating the correction at $Y$ from the change caused by the ADMM input. Starting from Eq.~\eqref{eq:app-spd-output-shift}, we obtain, pointwise,
\[
    \begin{aligned}
        \|D_\beta\|
        &=\bigl\|M\bigl[\beta d_t(C_\beta^\star)+e_\beta\bigr]\bigr\|\\
        &\le\beta\|M\|_{\mathrm{op}}\|d_t(C_\beta^\star)\|
          +\|M\|_{\mathrm{op}}\|e_\beta\|\\
        &\le\beta\|M\|_{\mathrm{op}}
          \bigl(\|d_t(Y)\|+\|d_t(C_\beta^\star)-d_t(Y)\|\bigr)+b_H\\
        &\le\beta\|M\|_{\mathrm{op}}
          \bigl(\|d_t(Y)\|+L\|C_\beta^\star-Y\|\bigr)+b_H\\
        &\le\beta\|M\|_{\mathrm{op}}
          \bigl(\|d_t(Y)\|+L\|B\|_{\mathrm{op}}\|D_\beta\|\bigr)+b_H\\
        &=\beta\|M\|_{\mathrm{op}}\|d_t(Y)\|
          +\delta_\beta\|D_\beta\|+b_H.
    \end{aligned}
\]
The first inequality uses the operator-norm bound and the triangle inequality. The second adds and subtracts $d_t(Y)$ and uses $\|e_\beta\|\le b$ with $b_H=\|M\|_{\mathrm{op}}b$. The next two inequalities use the $L$-Lipschitz property of $d_t$ and the identity $C_\beta^\star-Y=BD_\beta$, respectively. The final equality uses $\delta_\beta=\beta L\|M\|_{\mathrm{op}}\|B\|_{\mathrm{op}}$.

Since $\delta_\beta<1$, the term involving $D_\beta$ on the right can be moved to the left, bounding $D_\beta$ by a square-integrable random variable. Taking the $L^2$ norm and using $\|d_t(Y)\|_{L^2}=a_t$ gives
\begin{equation}
    \|D_\beta\|_{L^2}
    \le\frac{\beta\|M\|_{\mathrm{op}}a_t+b_H}{1-\delta_\beta}.
    \label{eq:app-spd-displacement}
\end{equation}
This proves square integrability of $A_\beta^\star$ and completes the justification of Eq.~\eqref{eq:app-risk-comparison-target}. The same Lipschitz estimate bounds the remainder as
\begin{equation}
    \begin{aligned}
        \|\xi_\beta\|_{L^2}
        &\le\delta_\beta\|D_\beta\|_{L^2}+b_H\\
        &\le\frac{\beta\|M\|_{\mathrm{op}}\delta_\beta}{1-\delta_\beta}a_t
          +\frac{b_H}{1-\delta_\beta}.
    \end{aligned}
    \label{eq:app-spd-remainder}
\end{equation}
We now use the decomposition $A_\beta^\star-\bar A(Y)=(I-\beta M)(Y-\bar A(Y))+\xi_\beta$ to turn the remainder bound into a bound on the corrected estimate's distance to the posterior mean. Expanding the steps gives
\begin{equation}
    \begin{aligned}
        \|A_\beta^\star-\bar A(Y)\|_{L^2}
        &=\bigl\|(I-\beta M)(Y-\bar A(Y))+\xi_\beta\bigr\|_{L^2}\\
        &\le\bigl\|(I-\beta M)(Y-\bar A(Y))\bigr\|_{L^2}
          +\|\xi_\beta\|_{L^2}\\
        &\le\|I-\beta M\|_{\mathrm{op}}\|Y-\bar A(Y)\|_{L^2}
          +\|\xi_\beta\|_{L^2}\\
        &=\|I-\beta M\|_{\mathrm{op}}a_t+\|\xi_\beta\|_{L^2}\\
        &\le\|I-\beta M\|_{\mathrm{op}}a_t
          +\frac{\beta\|M\|_{\mathrm{op}}\delta_\beta}{1-\delta_\beta}a_t
          +\frac{b_H}{1-\delta_\beta}\\
        &=\left(\|I-\beta M\|_{\mathrm{op}}
          +\frac{\beta\|M\|_{\mathrm{op}}\delta_\beta}{1-\delta_\beta}\right)a_t
          +\frac{b_H}{1-\delta_\beta}\\
        &=c_\beta a_t+\frac{b_H}{1-\delta_\beta}.
    \end{aligned}
    \label{eq:app-posterior-distance}
\end{equation}

We now substitute the two bounds into the risk difference stated at the start. The gain condition in Eq.~\eqref{eq:app-spd-gain-condition} ensures $a_t-b_H>c_\beta a_t+b_H/(1-\delta_\beta)\ge0$, so the squared lower bound strictly exceeds the squared upper bound. Consequently,
\begin{equation}
    \begin{aligned}
        &\mathbb E\|A_{\mathrm{off}}^\star-A_0\|^2
          -\mathbb E\|A_\beta^\star-A_0\|^2\\
        &\qquad=\mathbb E\|A_{\mathrm{off}}^\star-\bar A(Y)\|^2
          -\mathbb E\|A_\beta^\star-\bar A(Y)\|^2\\
        &\qquad\ge(a_t-b_H)^2
          -\left(c_\beta a_t+\frac{b_H}{1-\delta_\beta}\right)^2>0.
    \end{aligned}
    \label{eq:app-risk-gain}
\end{equation}
This proves Eq.~\eqref{eq:gdsm-risk-improvement}: under the stated gain condition, the corrected limit is closer to the posterior mean in mean square and therefore has strictly lower expected reconstruction error. Together with Part 2, this establishes convergence and the claimed error improvement over the same solver with the score correction disabled.
\end{proof}

\clearpage
\section{Implementation Details}
\label{app:implementation-details}
\label{app:training-details}

\subsection{Task formulations and analytical ADMM}
\label{app:analytical-solvers}

All three tasks follow the ADMM formulation in Eq.~\eqref{eq:admm}, with the common denoiser input $C^k=A^{k+1}+U^k$ and task-specific observation models and analytical prior updates. We detail these components and their implementation below. For matched comparisons, we disable the learned corrections while keeping the data update, analytical prior, and postprocessing fixed.

\noindent\textbf{Reference-guided HSI restoration.} We represent the hyperspectral image as $A=EB+\mu$, where $B$ contains the spectral coefficients, and the spectral basis $E$ and mean $\mu$ are estimated from the observations and kept fixed. Let $\mathcal H$ denote spatial blur and downsampling, and let $\mathcal R_{\mathrm{PAN}}$ denote the PAN spectral response. The data-fidelity term combines consistency with the low-resolution HSI and the PAN image: \begin{equation} f(A)=\frac12\|\mathcal H(A)-Y_{\mathrm{LR}}\|_{Q(W)}^2+\frac12\|\mathcal R_{\mathrm{PAN}}(A)-Y_{\mathrm{PAN}}\|_2^2, \label{eq:app-hipandas-fidelity} \end{equation} where $\|r\|_{Q(W)}^2=\langle r,Q(W)r\rangle$, and $Q(W)$ incorporates observation-derived reliability weights. We implement the analytical ADMM solver in coefficient coordinates, with auxiliary variable $V_B$ and scaled dual variable $U_B$: \begin{equation} \begin{aligned} B^{k+1}&=\arg\min_B f(EB+\mu)+\frac{\rho}{2}\|B-V_B^k+U_B^k\|_F^2,\\ C_B^k&=B^{k+1}+U_B^k,\\ V_B^{k+1}&=\mathcal T_\tau(C_B^k),\\ U_B^{k+1}&=U_B^k+B^{k+1}-V_B^{k+1}. \end{aligned} \label{eq:app-hipandas-admm} \end{equation} The data update is a quadratic subproblem solved by preconditioned conjugate gradients. The analytical map $\mathcal T_\tau$ applies local singular-value soft thresholding to promote low-rank structure in the coefficient patches. Our learned correction modifies $C_B^k$ before this analytical update, following the prior interface defined in the main text.

\noindent\textbf{Single-image HSI denoising.} We represent the hyperspectral image as $A=EB$, where $E$ is a fixed orthonormal spectral basis estimated from the noisy observations and $B$ contains the spectral coefficients. For mixed noise, we additionally introduce a sparse component $S$ and a column component $L$, with the latter constant along the image height. The weighted data-fidelity term is \begin{equation} f(A;S,L)=\frac12\|W^{1/2}\odot(Y-A-S-L)\|_F^2, \label{eq:app-hsi-fidelity} \end{equation} where $Y$ is the noisy HSI and $W$ contains observation-derived reliability weights. Gaussian denoising sets $S=L=0$. We initialize $V_B^0=E^\top Y$, $U_B^0=0$, and $S^0=L^0=0$. Holding the noise estimates fixed during the coefficient update, the analytical ADMM steps are \begin{equation} \begin{aligned} B^{k+1}&=\arg\min_B f(EB;S^k,L^k)+\frac{\rho}{2}\|B-V_B^k+U_B^k\|_F^2,\\ C_B^k&=B^{k+1}+U_B^k,\\ V_B^{k+1}&=\mathcal T_{\mathrm{HSI}}(C_B^k),\\ U_B^{k+1}&=U_B^k+B^{k+1}-V_B^{k+1}. \end{aligned} \label{eq:app-hsi-admm} \end{equation} The coefficient data update is obtained by solving a positive-definite linear system at each pixel. In the LocalSVT/TV configuration, $\mathcal T_{\mathrm{HSI}}$ applies local singular-value soft thresholding followed by a finite-step second-order total variation (TV) update. DCT/TV configurations instead use a joint transform-sparsity and TV update. For mixed noise, each ADMM iteration is followed by robust updates using the residual $T^{k+1}=Y-EV_B^{k+1}$: the column component is estimated by soft-thresholding the height-wise medians of $T^{k+1}-S^k$ on the detected column support, with column centering where configured, and the sparse component is updated as $S^{k+1}=\operatorname{soft}(T^{k+1}-L^{k+1},\tau_{\mathrm{imp}})$, where $\operatorname{soft}$ denotes element-wise soft thresholding. These noise estimates enter the next coefficient data update. The restored image is $\widehat A=EV_B^K$. Our learned correction modifies $C_B^k$ before the analytical prior update, following the interface defined in the main text.

\noindent\textbf{MRI denoising.}
We restore the magnitude image $A$ directly from the noisy observation $Y$. The scaled Rician data-fidelity term is
\begin{equation}
    f_{\mathrm R}(A)
    =\sum_p\left[
        \frac12 A_p^2
        -\widehat\sigma_n^2
        \log I_0\!\left(\frac{Y_pA_p}{\widehat\sigma_n^2}\right)
    \right],
    \qquad A\geq0,
    \label{eq:app-mri-fidelity}
\end{equation}
where $p$ indexes pixels, $\widehat\sigma_n$ is an observation-derived noise estimate, and $I_0$ is the modified Bessel function of the first kind of order zero. Terms independent of $A$ are omitted. For multicoil M4Raw magnitude images, this data term serves as a likelihood surrogate.

We initialize $V^0=Y$ and $U^0=0$. With the nonnegative data-step center $b^k=\max(V^k-U^k,0)$ used in our implementation, the analytical ADMM updates are
\begin{equation}
    \begin{aligned}
        A^{k+1}
        &=\arg\min_{A\geq0}
        f_{\mathrm R}(A)
        +\frac{\rho}{2}\|A-b^k\|_F^2,\\
        C^k&=A^{k+1}+U^k,\\
        V^{k+1}&=\mathcal T_{\mathrm{MRI}}(C^k),\\
        U^{k+1}&=U^k+A^{k+1}-V^{k+1}.
    \end{aligned}
    \label{eq:app-mri-admm}
\end{equation}
The data subproblem is approximately solved by fixed-point/MM iterations. The analytical map $\mathcal T_{\mathrm{MRI}}$ combines sparsity of the non-DC coefficients of the two-dimensional DCT with second-order total variation (TV), promoting transform-domain sparsity and spatial regularity. This joint prior update is approximately computed by primal--dual iterations. Our learned correction modifies $C^k$ before the analytical prior update, following the interface defined in the main text.

\subsection{Network architecture and state-bank construction}
\label{app:network-details}
\label{app:bank-details}

\noindent\textbf{Correction network.}
The correction network is a shallow conditional U-Net with feature widths 48 and 96, four residual blocks, and one downsampling--upsampling pair. Each residual block uses GroupNorm and SiLU activations and is conditioned on a 128-dimensional embedding of the perturbation level. The input concatenates the normalized state, a one-channel guide, and three scalar condition maps. The final convolution is zero-initialized. The state has $r=6$, $12$, and $1$ channels for reference-guided HSI restoration, single-image HSI denoising, and MRI denoising, respectively. HiPandas uses the observed PAN image as guidance; HSI denoising and MRI use guidance derived from their noisy observations. Each network contains approximately $0.64$--$0.65$ million trainable parameters, with the architecture and exact counts given in Table~\ref{tab:network-architecture}. Dual-track inference uses two networks with the same architecture and independent weights, doubling the parameter count.

\begin{table}[htbp]
    \centering
    \small
    \setlength{\tabcolsep}{4pt}
    \renewcommand{\arraystretch}{1.12}
    \caption{Architecture of one correction network. $r$ is the number of state channels, and $H\times W$ is the input spatial size. Parameter counts include all trainable weights and biases.}
    \label{tab:network-architecture}
    \begin{tabularx}{\linewidth}{@{}lXcr@{}}
        \toprule
        Stage & Operation & Output size & Parameters \\
        \midrule
        Input & State, guide, and condition maps & $(r+4)\times H\times W$ & -- \\
        Conditioning & MLP: $129\to128\to128$ & $128$ & 33,152 \\
        Input projection & $3\times3$ convolution & $48\times H\times W$ & $432r+1{,}776$ \\
        Encoder block 1 & Residual block & $48\times H\times W$ & 47,952 \\
        Downsampling & $4\times4$ convolution, stride 2 & $96\times\frac H2\times\frac W2$ & 73,824 \\
        Encoder block 2 & Residual block & $96\times\frac H2\times\frac W2$ & 178,848 \\
        Bottleneck & Residual block & $96\times\frac H2\times\frac W2$ & 178,848 \\
        Upsampling & $4\times4$ transposed convolution, stride 2 & $48\times H\times W$ & 73,776 \\
        Skip connection & Add encoder block 1 output & $48\times H\times W$ & -- \\
        Decoder block & Residual block & $48\times H\times W$ & 47,952 \\
        Output projection & GroupNorm, SiLU, $3\times3$ convolution & $r\times H\times W$ & $433r+96$ \\
        \midrule
        \multicolumn{3}{l}{Total: reference-guided HSI restoration ($r=6$)} & \textbf{0.641M} \\
        \multicolumn{3}{l}{Total: single-image HSI denoising ($r=12$)} & \textbf{0.647M} \\
        \multicolumn{3}{l}{Total: MRI denoising ($r=1$)} & \textbf{0.637M} \\
        \bottomrule
    \end{tabularx}
\end{table}
\noindent\textbf{Training states.}
We first run the analytical ADMM solver and collect its denoiser inputs into a state bank $\mathcal B_0$. For single-track learning, the correction network is trained directly on $\mathcal B_0$. For dual-track learning, we first train the structure network on $\mathcal B_0$ and freeze its weights. We then run ADMM with the structure correction enabled to construct a second bank $\mathcal B_1$, on which the complementary network is trained. This sequential procedure allows the complementary network to adapt to states generated after the structure correction. Each network is trained on a fixed bank without backpropagating through the ADMM iterations, and all network weights are fixed during final inference.

\noindent\textbf{Fixed normalization.}
For each bank and channel, we compute a median center $m_i$ and scale
\begin{equation}
    s_i=\max\!\left\{\frac{q_{0.99}-q_{0.01}}{2},10^{-4}\right\},
    \qquad \mathcal S_i(C)=(C-m_i)\oslash s_i.
    \label{eq:app-bank-normalization}
\end{equation}
The statistics are fixed before training and reused at inference. In dual-track learning, the two banks may have different normalization statistics; branch outputs are therefore restored to the original coefficient scale before projection.

\subsection{Self-supervised objectives and optimization}
\label{app:training-objectives}
\label{app:bounded-training}

\noindent\textbf{Score regression.}
For a normalized bank state $Z$, we sample $\epsilon\sim\mathcal N(0,I)$ and construct
\begin{equation}
    Z_t=Z+\sqrt{\Delta_t}\epsilon,
    \qquad
    \Delta_t=\nu^2(\sigma_t^2-1),
    \qquad
    \kappa_t=\nu^2\sigma_t^2,
    \label{eq:app-perturbation-parameters}
\end{equation}
where $\nu>0$ controls the artificial perturbation scale and $\sigma_t>1$ specifies its level.
We train the predicted correction $D_i(Z_t,t)=\kappa_t s_{\theta_i}(Z_t,t)$ to match the observable target $\kappa_t(Z-Z_t)/\Delta_t$.
Before output bounding and branch projection, the regression objective is
\begin{equation}
    \mathcal L_{\mathrm{score},i}
    =
    \mathbb E\!\left[
        w_i(t)
        \left\|
            D_i(Z_t,t)-\kappa_t\Delta_t^{-1}(Z-Z_t)
        \right\|_F^2
    \right],
    \qquad
    w_i(t)=(N_i\gamma_i\sigma_t^2)^{-1}.
    \label{eq:orth-bounded-gdsm}
\end{equation}
Here $N_i$ is the number of tensor entries per training sample, and $\gamma_i=0.01a_i^{-1}$ incorporates the external score-loss coefficient $a_i$.
When output bounding or branch projection is used, the same processing is applied to both predictions and targets before computing the squared error.

\noindent\textbf{Auxiliary losses.}
When a held-out observation loss is used, a fixed set $\Omega$ is masked when constructing the state bank and guidance.
The original noisy measurements at these locations, denoted by $Y_\Omega$, serve as training targets.
Let $\widehat Y_{\Omega,i}=[\mathcal H(\widehat A_i)]_\Omega$ denote the corresponding predicted measurements.
For reference-guided HSI restoration, we minimize their weighted absolute difference:
\begin{equation}
    \mathcal L_{\mathrm{held},i}
    =
    |\Omega|^{-1}
    \mathbb E\!\left[
        \left\|
            w_\Omega\odot
            (\widehat Y_{\Omega,i}-Y_\Omega)
        \right\|_1
    \right],
    \label{eq:app-held-loss}
\end{equation}
where $w_\Omega$ contains the observation-reliability weights.
For MRI, we instead average the scaled Rician discrepancy in Eq.~\eqref{eq:app-mri-fidelity} over the same held-out locations. The auxiliary objective is
\begin{equation}
    \mathcal L_{\mathrm{aux},i}
    =\mathcal L_{\mathrm{held},i}
    +\omega_{\mathrm p,i}\mathcal L_{\mathrm{PAN},i}
    \label{eq:app-aux-loss}
\end{equation}
The PAN term is $\mathcal L_{\mathrm{PAN},i}=\mathbb E\operatorname{mean}|\mathcal R_{\mathrm{PAN}}(\widehat A_i)-Y_{\mathrm{PAN}}|$ and is used only for reference-guided HSI restoration. 

\noindent\textbf{Total objective.}
Consistent with the main text, each network is trained using
\begin{equation}
    \mathcal L_i
    =\mathcal L_{\mathrm{score},i}
    +\lambda\mathcal L_{\mathrm{aux},i}.
    \label{eq:orth-total-loss}
\end{equation}
The weights depend on the task and training configuration. 

\noindent\textbf{Optimization.}
HiPandas and MRI use AdamW with learning rate $10^{-4}$, zero weight decay, batch size four, gradient-norm clipping at one, and an EMA decay of $0.999$. Inference uses frozen EMA weights. MRI trains for 5,000 updates per network. 

\subsection{Projectors and inference settings}
\label{app:projector-construction}
\label{app:correction-details}

We construct two complementary orthogonal projectors to specify the correction subspace of each branch. The structure projector $P_s$ selects directions associated with observation-supported column interference, while $P_g$ selects their orthogonal complement. Under vectorization of the coefficient maps, let $Q$ contain an orthonormal basis of the structural subspace. The projectors are defined as
\begin{equation}
    P_s=QQ^\top,
    \qquad
    P_g=I-P_s,
    \qquad
    Q^\top Q=I.
    \label{eq:app-structural-projectors}
\end{equation}
They satisfy $P_s+P_g=I$ and $P_sP_g=0$, so their ranges are complementary and mutually orthogonal.

For HSI column artifacts, the structural subspace combines spectral support with low-frequency variation along the image height. For each spatial column $j$, let $M_j$ be a diagonal mask indicating the spectral bands with detected interference support. The eigenvectors associated with the numerically nonzero eigenvalues of $E^\top M_jE$ form an orthonormal basis $Q_j$ in the spectral coefficient coordinates. We combine these directions with the first $q$ orthonormal height-wise DCT modes to construct the structural subspace independently for each column. This construction captures supported patterns that are constant or slowly varying along the column.

Both networks receive the complete ADMM denoiser input. Their corrections are rescaled to the original coefficient coordinates and projected by $P_s$ and $P_g$, respectively, before entering the analytical prior update. The structural subspace may also contain image signal, so the network learns the correction within the selected directions. When reliable structural support is unavailable, we use $P_s=0$ and $P_g=I$.

\subsection{Datasets and Implementation Details}
\label{app:experimental-settings}
\label{app:datasets-evaluation}

\noindent\textbf{Reference-guided HSI restoration.}
\label{app:hipandas-data}
We use the Dongying dataset\footnote{\url{https://github.com/shuangxu96/Hipandas}} from HiPandas \citep{xu2025hipandas}. It contains 81 samples with 32 spectral bands. Each sample consists of a noisy $64\times64\times32$ low-resolution HSI and a $256\times256$ PAN image, with a target HSI of size $256\times256\times32$. The spatial scale factor is four.
The Gaussian conditions include G10 and G30, as well as GNI
with band-dependent, non-i.i.d. Gaussian noise. The mixed conditions, Mix35, Mix55, and Mix55-G30, combine Gaussian noise with stripe and bad-line interference at different severity levels.
For each condition, all 81 noisy samples contribute states to a shared training bank, and the resulting network or network pair is shared across the samples. Training uses $64\times64$ crops in high-resolution state coordinates, corresponding to $16\times16$ low-resolution observation regions. Inference restores each complete sample. Clean high-resolution images are excluded from the training losses and are used for parameter selection and evaluation. Metrics are computed per sample and then averaged over all 81 samples. 

\noindent\textbf{Single-image HSI denoising.}
\label{app:hsi-data}
We use Qingpu\footnote{\url{https://www.3sobs.cn/qingpu-hsi-dataset}} and Xiong'an\footnote{\url{https://doi.org/10.3974/geodb.2021.01.02.V1}}, with one noisy HSI as the input to each restoration problem. The experimental scenes have spatial dimensions of $1580\times3750$, with 251 and 256 spectral bands for Qingpu and Xiong'an, respectively.
Both datasets are evaluated under Gaussian conditions G30/G50/G70 and mixed conditions M30/M50/M70.
The mixed conditions combine band-dependent Gaussian noise with stripe noise, dead lines, and salt-and-pepper noise.
The scenes are divided into $256\times256$ spatial patches, which are restored independently using a network shared within each scene and noise condition.
Training uses crops of solver states generated from the noisy observations.

\noindent\textbf{MRI denoising.}
\label{app:mri-data}
We use BrainWeb\footnote{\url{https://brainweb.bic.mni.mcgill.ca/selection_normal.html}} and M4Raw\footnote{\url{https://github.com/mylyu/M4Raw}} to evaluate denoising under simulated and acquired noise. BrainWeb provides T1, T2, and PD images at 1~mm spatial resolution, with a native axial image size of $217\times181$ pixels. We use the official 3\%, 5\%, and 7\% noise levels, yielding nine conditions, and evaluate 34 slices per condition, giving 306 slice--condition pairs. For M4Raw, we evaluate all 25 test subjects, including the three used for parameter selection. Each subject provides 18 slices of size $256\times256$ for each of the T1, T2, and FLAIR contrasts, totaling 1,350 slices. We use the acquired noisy images without additional corruption. References are averages of repeated acquisitions that include the input repetition and therefore remain noisy and correlated with the input.

\noindent\textbf{HiPandas parameters.}
Table~\ref{tab:app-hipandas-parameters} lists the HiPandas parameters.
$K$ is the number of outer ADMM iterations and $\rho$ is the ADMM penalty.
$\tau$ is the singular-value soft-threshold of the local low-rank prior.
$d$ is the componentwise correction cap in normalized state coordinates before gain scaling.
$\beta_s$ and $\beta_g$ are the nominal gains of the structure and
full-space/complementary corrections, respectively.
$N_s$ and $N_g$ are the optimizer update counts of the corresponding networks;
-- indicates an unused network.
Gaussian and GNI conditions use Full-space inference; mixed conditions use Structure-guided inference.

\begin{table}[!htbp]
    \centering
    \small
    \setlength{\tabcolsep}{3.5pt}
    \renewcommand{\arraystretch}{1.08}
    \caption{HiPandas solver and correction parameters.}
    \label{tab:app-hipandas-parameters}
        \begin{tabular}{@{}l*{8}{r}@{}}
        \toprule
        Condition & $K$ & $\rho$ & $\tau$ & $d$ & $\beta_s$ & $\beta_g$ & $N_s$ & $N_g$ \\
        \midrule
        G10 & 4 & 1 & 0.105 & 0.10 & 0 & 1 & -- & 5000 \\
        G30 & 8 & 8 & 0.105 & 0.10 & 0 & 1 & -- & 5000 \\
        GNI & 8 & 4 & 0.105 & 0.10 & 0 & 1 & -- & 5000 \\
        Mix35     & 8 & 8 & 0.070 & 0.10 & 0.75 & 1.25 & 10000 & 5000 \\
        Mix55     & 8 & 4 & 0.070 & 0.10 & 0.75 & 1.25 & 5000  & 5000 \\
        Mix55-G30 & 8 & 4 & 0.070 & 0.10 & 0.50 & 0.50 & 5000  & 5000 \\
        \bottomrule
    \end{tabular}
\end{table}

\noindent\textbf{HSI denoising parameters.}
Table~\ref{tab:app-hsi-parameters} lists the parameters for Qingpu and Xiong'an,
using the same notation as above.
Gaussian conditions use Full-space inference; mixed conditions use Structure-guided inference.

\begin{table}[!htbp]
    \centering
    \small
    \setlength{\tabcolsep}{3pt}
    \renewcommand{\arraystretch}{1.08}
    \caption{HSI denoising solver and correction parameters.}
    \label{tab:app-hsi-parameters}
    \begin{tabular}{@{}l*{7}{r}@{}}
        \toprule
        Condition & $K$ & $\rho$ & $d$ & $\beta_s$ & $\beta_g$ & $N_s$ & $N_g$ \\
        \midrule
        Qingpu G30 & 30 & 2 & 0.1 & 0 & 0.5 & -- & 5000 \\
        Qingpu G50 & 30 & 2 & 0.1 & 0 & 0.75 & -- & 6000 \\
        Qingpu G70 & 30 & 2 & 0.1 & 0 & 0.75 & -- & 5000 \\
        Qingpu M30 & 30 & 2 & 0.1 & 0.125 & 0.5 & 2000 & 2000 \\
        Qingpu M50 & 30 & 2 & 0.1 & 0.375 & 0.75 & 3000 & 3000 \\
        Qingpu M70 & 30 & 2 & 0.1 & 0.1875 & 0.75 & 2000 & 2000 \\
        Xiong'an G30 & 30 & 2 & 0.1 & 0 & 0.5 & -- & 4000 \\
        Xiong'an G50 & 30 & 2 & 0.1 & 0 & 0.5 & -- & 4000 \\
        Xiong'an G70 & 30 & 2 & 0.1 & 0 & 0.75 & -- & 6000 \\
        Xiong'an M30 & 30 & 2 & 0.1 & 0.125 & 0.5 & 2000 & 2000 \\
        Xiong'an M50 & 30 & 2 & 0.1 & 0.75 & 0.75 & 2500 & 2500 \\
        Xiong'an M70 & 30 & 2 & 0.1 & 0.1875 & 0.75 & 2000 & 2000 \\
        \bottomrule
    \end{tabular}
\end{table}

\noindent\textbf{MRI parameters.}
Tables~\ref{tab:app-brainweb-parameters} and~\ref{tab:app-m4raw-parameters}
list the parameters for BrainWeb and M4Raw, respectively.
Both datasets use Full-space inference.
$K$, $\rho$, and $d$ have the same meanings as above;
$N_g$ is the number of optimizer updates for the full-space network.
$c$ and $a$ are the noise-scaled prior coefficients for DCT sparsity
and TGV regularization, respectively, and $J$ is the number of inner
iterations used to evaluate their joint proximal map.
$\gamma_\sigma$ multiplies the estimated noise standard deviation before
it is used by the solver.
$\beta$ is the nominal correction gain, and $\eta_k$ is its iteration-dependent
multiplier, so the applied gain is $\beta\eta_k$ at outer iteration
$k\in\{0,\ldots,K-1\}$.
In Table~\ref{tab:app-m4raw-parameters}, an arrow denotes the training value followed by the
inference value; a single value is shared by both stages.
$\mathbf{1}_{\{k\geq10\}}$ activates the correction during the last 20 of
30 iterations, and $[x]_0^1=\min\{1,\max\{0,x\}\}$ defines the ramp schedule.

\begin{table}[!htbp]
    \centering
    \small
    \setlength{\tabcolsep}{3.5pt}
    \renewcommand{\arraystretch}{1.08}
    \caption{BrainWeb solver and correction parameters.}
    \label{tab:app-brainweb-parameters}
        \begin{tabular}{@{}l*{10}{r}@{}}
        \toprule
        Condition & $K$ & $\rho$ & $c$ & $a$ & $J$ & $\gamma_\sigma$ & $d$ & $\beta$ & $N_g$ & $\eta_k$ \\
        \midrule
        T1, 3\% & 30 & 2 & 0.1875 & 0.20 & 20 & 1 & 0.10 & 0.25 & 5000 & 1 \\
        T1, 5\% & 30 & 2 & 0.1875 & 0.20 & 20 & 1 & 0.10 & 0.50 & 5000 & 1 \\
        T1, 7\% & 30 & 2 & 0.1875 & 0.20 & 20 & 1 & 0.10 & 0.50 & 5000 & 1 \\
        T2, 3\% & 30 & 2 & 0.1875 & 0.20 & 20 & 1 & 0.10 & 0.25 & 5000 & 1 \\
        T2, 5\% & 30 & 2 & 0.1875 & 0.20 & 20 & 1 & 0.10 & 0.50 & 5000 & 1 \\
        T2, 7\% & 30 & 2 & 0.1875 & 0.20 & 20 & 1 & 0.10 & 0.75 & 5000 & 1 \\
        PD, 3\% & 30 & 2 & 0.1875 & 0.20 & 20 & 1 & 0.10 & 0.25 & 5000 & 1 \\
        PD, 5\% & 30 & 2 & 0.1875 & 0.20 & 20 & 1 & 0.10 & 0.50 & 5000 & 1 \\
        PD, 7\% & 30 & 2 & 0.1875 & 0.20 & 20 & 1 & 0.10 & 0.75 & 5000 & 1 \\
        \bottomrule
    \end{tabular}
\end{table}

\begin{table}[t]
    \centering
    \small
    \setlength{\tabcolsep}{3pt}
    \renewcommand{\arraystretch}{1.08}
    \caption{M4Raw solver and correction parameters.}
    \label{tab:app-m4raw-parameters}
    \resizebox{\linewidth}{!}{%
    \begin{tabular}{@{}lrrccrrrrrc@{}}
        \toprule
        Contrast & $K$ & $\rho$ & $c$ & $a$ & $J$ & $\gamma_\sigma$ & $d$ & $\beta$ & $N_g$ & $\eta_k$ \\
        \midrule
        T1 & 30 & 2 & $0.03125\to0.015625$ & $0.350\to0.175$ & 20 & 1.25 & 0.20 & 1.00 & 5000 & $[(k-15)/15]_0^1$ \\
        T2 & 30 & 2 & 0.09375 & 0.175 & 20 & 1.25 & 0.10 & 0.50 & 5000 & $\mathbf{1}_{\{k\geq10\}}$ \\
        FLAIR & 30 & 2 & 0.125 & 0.050 & 20 & 1.25 & 0.10 & 0.50 & 5000 & 1 \\
        \bottomrule
    \end{tabular}%
    }
\end{table}

\begingroup
\makeatletter
\setlength{\@fptop}{0pt}
\makeatother
\clearpage
\endgroup

\clearpage
\section{Additional Experimental Results}
\label{app:additional-results}

\subsection{Controlled Experiments for the Theoretical Analysis}
\label{app:theory-experiment-details}

\noindent\textbf{Experimental setup.}
We use $16\times16$, $32\times32$, and $64\times64$ Gaussian states with a known ideal score. Normalization is the identity. For $d=m^2$, the clean covariance $\Sigma$ is diagonal with entries $0.2$, $0.5$, $1$, and $2$ repeated equally; independent Gaussian corruptions have total variance $\kappa_t=0.5$. The data fidelity is $\frac12\langle A-Y,H(A-Y)\rangle$, where $H$ has eigenvalues $0.25$, $0.5$, $2$, and $4$, repeated equally in a fixed orthogonal basis, and $\rho=1$. The analytical map is exact singular-value soft thresholding on the complete matrix, with threshold $\tau=\gamma a_t/\sqrt m$, where $a_t^2=\kappa_t^2\operatorname{tr}((\Sigma+\kappa_t I)^{-1})$. Thus, $\gamma=b/a_t$ measures relative analytical-prior strength. Corrected and correction-disabled solvers share observations, initialization, data fidelity, prior, and penalty.

\begin{figure}[!htbp]
    \centering
    \includegraphics[width=\linewidth]{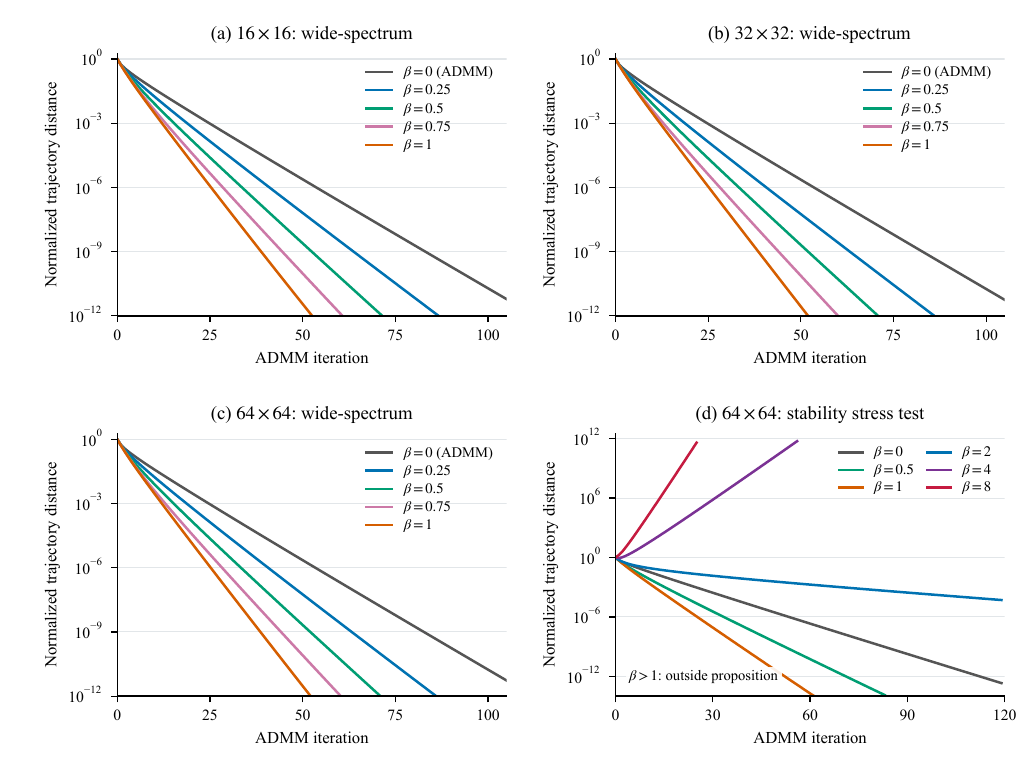}
    \caption{Wide-spectrum convergence and stability. (a)--(c) Original ADMM trajectories across dimensions at $\gamma=0.15$. (d) Additional correction strengths at $64\times64$; $\beta>1$ lies outside the theorem's range. Growing runs stop at a prescribed numerical safety threshold. Panel (d) uses wider axis ranges.}
    \label{fig:app-theory-convergence}
\end{figure}

\noindent\textbf{Convergence and stability.}
At $\gamma=0.15$, we run original ADMM on 16 observations from two different initializations and measure the root-mean-square distance between their prior-input trajectories, normalized by its initial value. Figures~\ref{fig:app-theory-convergence}(a)--(c) compare five correction strengths across dimensions; larger tested corrections yield faster decay in this setting. For this Hessian, the theorem's sufficient contraction condition holds for $\beta<7/15$. The tested larger corrections within $[0,1]$ also converge, demonstrating the conservatism of that condition. Panel (d) extends the sweep beyond the theorem's range: $\beta=4$ and $8$ produce growing trajectory distances and primal residuals, whereas $\beta=2$ continues to decrease during the recorded iterations. Failure of the sufficient condition alone does not imply divergence.

\begin{figure}[!htbp]
    \centering
    \includegraphics[width=\linewidth]{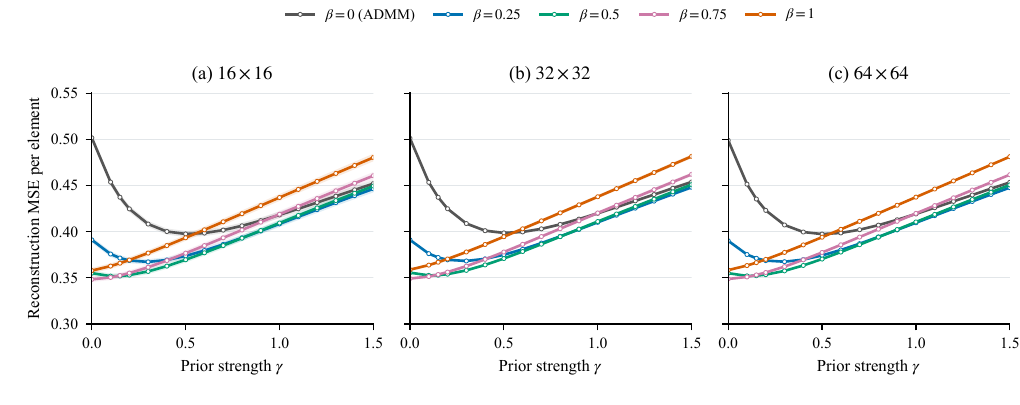}
    \caption{Reconstruction error under wide-spectrum fidelity. Each panel fixes the state dimension; lower curves indicate better restoration. The gray curve is analytical ADMM ($\beta=0$). Markers denote evaluated prior strengths, connecting segments are visual guides, and shading shows pointwise 95\% intervals computed across independent test matrices. All panels share the same vertical scale.}
    \label{fig:app-theory-risk}
\end{figure}

\noindent\textbf{Reconstruction error.}
We evaluate 512 independent test matrices per dimension, reusing the same clean signals and noisy observations across parameter settings. Figure~\ref{fig:app-theory-risk} plots per-element reconstruction error against $\gamma$ for $\beta\in\{0,0.25,0.5,0.75,1\}$. The 17 evaluated prior strengths span $0$ to $1.5$ in increments of $0.1$, with $0.15$ additionally included. The main text shows the $64\times64$ case, using the same data and correction strengths. For each setting, we report
\begin{equation}
    \widehat R_d(\beta,\gamma)
    =\frac{1}{Nd}\sum_{j=1}^{N}
      \|A_{\beta,j}^\star-A_{0,j}\|_F^2,
    \label{eq:app-theory-empirical-risk}
\end{equation}
where $A_{\beta,j}^\star$ is computed at prior strength $\gamma$, and $\beta=0$ gives analytical ADMM. Lower values indicate better reconstruction. The error curves illustrate the interaction between analytical regularization and score correction: stronger correction is not uniformly better. The tested nonzero correction strengths lie outside the sufficient risk-improvement region, so these comparisons characterize empirical performance beyond the certified settings.

Final risks are computed with Anderson acceleration of the ADMM state map, checking both original fixed-point and primal residuals below $10^{-10}$. Selected estimates are also checked against direct ADMM. All convergence curves use original ADMM iterations.

\providecommand{\methodname}{S\textsuperscript{3}-PnP}

\subsection{Additional quantitative results}
\label{app:additional-restoration-quantitative}
\noindent\textbf{Reference-guided HSI restoration.}
Tables~\ref{tab:app-hipandas-gaussian} and~\ref{tab:app-hipandas-mixed}
report G10/G30/GNI and mixed conditions, respectively, with 81 patches per condition.
Bold and underlined entries indicate the best and second-best distinct values in each column.

\begin{table}[!htbp]
\centering\small
\setlength{\tabcolsep}{3pt}
\renewcommand{\arraystretch}{1.08}
\caption{Reference-guided HSI restoration on Dongying under Gaussian/GNI noise.}
\label{tab:app-hipandas-gaussian}
\resizebox{\linewidth}{!}{%
\begin{tabular}{@{}c*{12}{c}@{}}
\toprule
 & \multicolumn{4}{c}{G10} & \multicolumn{4}{c}{G30} & \multicolumn{4}{c}{GNI} \\
\cmidrule(lr){2-5}\cmidrule(lr){6-9}\cmidrule(lr){10-13}
Method & PSNR $\uparrow$ & SSIM $\uparrow$ & SAM $\downarrow$ & ERGAS $\downarrow$ & PSNR $\uparrow$ & SSIM $\uparrow$ & SAM $\downarrow$ & ERGAS $\downarrow$ & PSNR $\uparrow$ & SSIM $\uparrow$ & SAM $\downarrow$ & ERGAS $\downarrow$ \\
\midrule
PLRDiff + HIR-Diff & 34.52 & 0.8215 & 6.73 & 16.49 & 32.20 & 0.8115 & 8.96 & 20.36 & 31.01 & 0.7670 & 12.73 & 23.66 \\
PLRDiff + PWRCTV & 38.10 & 0.9445 & 5.42 & 12.22 & 33.43 & 0.9156 & 9.02 & 18.93 & 32.73 & 0.9082 & 12.05 & 21.31 \\
SwinIR + HIR-Diff & 33.88 & 0.8512 & 4.69 & 22.02 & 27.79 & 0.4978 & 10.21 & 33.18 & 27.18 & 0.4962 & 12.66 & 35.11 \\
SwinIR + PWRCTV & 31.66 & 0.7815 & 8.33 & 65.58 & 21.16 & 0.3662 & 29.96 & 422.81 & 22.62 & 0.4464 & 31.18 & 489.23 \\
R-PNN + HIR-Diff & 33.80 & 0.8068 & 6.22 & 17.49 & 29.10 & 0.6743 & 9.28 & 25.03 & 29.59 & 0.6866 & 10.90 & 25.12 \\
R-PNN + PWRCTV & 36.71 & 0.9240 & 5.01 & 13.53 & 29.58 & 0.6375 & 14.46 & 28.57 & 30.96 & 0.7174 & 15.31 & 29.07 \\
HIR-Diff + PLRDiff & 33.48 & 0.7606 & 8.77 & 42.15 & 27.52 & 0.5859 & 17.65 & 338.12 & 23.06 & 0.3942 & 31.89 & 467.35 \\
HIR-Diff + SwinIR & 36.23 & 0.8612 & 3.88 & 13.07 & 31.00 & 0.6963 & 7.59 & 21.15 & 30.03 & 0.6716 & 9.53 & 24.05 \\
HIR-Diff + R-PNN & 37.90 & 0.9390 & 4.12 & 11.52 & 34.82 & 0.9157 & 6.08 & 15.97 & 33.25 & 0.9048 & 8.04 & 18.57 \\
PWRCTV + PLRDiff & 38.52 & 0.9116 & 3.88 & 11.28 & 35.46 & 0.8691 & 5.97 & 30.24 & 34.41 & 0.8217 & 8.84 & 23.09 \\
PWRCTV + SwinIR & 37.50 & 0.9080 & 3.10 & 11.33 & 34.54 & 0.8638 & 5.12 & 15.96 & 34.60 & 0.8602 & 5.76 & 16.66 \\
UHiPandas & \underline{40.54} & \underline{0.9543} & \underline{3.04} & \underline{8.52} & \underline{36.82} & \underline{0.9261} & \underline{4.66} & \underline{12.34} & \underline{37.35} & \underline{0.9300} & \underline{4.46} & \underline{11.74} \\
\midrule
\methodname{} (Ours) & \textbf{41.91} & \textbf{0.9614} & \textbf{2.58} & \textbf{6.89} & \textbf{38.52} & \textbf{0.9428} & \textbf{3.97} & \textbf{10.57} & \textbf{38.84} & \textbf{0.9448} & \textbf{3.80} & \textbf{10.13} \\
\bottomrule
\end{tabular}}
\end{table}

\begin{table}[!htbp]
\centering\small
\setlength{\tabcolsep}{3pt}
\renewcommand{\arraystretch}{1.08}
\caption{Reference-guided HSI restoration on Dongying under mixed noise.}
\label{tab:app-hipandas-mixed}
\resizebox{\linewidth}{!}{%
\begin{tabular}{@{}c*{12}{c}@{}}
\toprule
 & \multicolumn{4}{c}{Mix35} & \multicolumn{4}{c}{Mix55} & \multicolumn{4}{c}{Mix55-G30} \\
\cmidrule(lr){2-5}\cmidrule(lr){6-9}\cmidrule(lr){10-13}
Method & PSNR $\uparrow$ & SSIM $\uparrow$ & SAM $\downarrow$ & ERGAS $\downarrow$ & PSNR $\uparrow$ & SSIM $\uparrow$ & SAM $\downarrow$ & ERGAS $\downarrow$ & PSNR $\uparrow$ & SSIM $\uparrow$ & SAM $\downarrow$ & ERGAS $\downarrow$ \\
\midrule
PLRDiff + HIR-Diff & 30.90 & 0.7649 & 12.30 & 26.04 & 30.70 & 0.7687 & 13.35 & 35.69 & 31.63 & 0.7997 & 11.03 & 34.60 \\
PLRDiff + PWRCTV & 32.58 & 0.9079 & 11.79 & 23.82 & 32.13 & 0.9000 & 12.92 & 34.07 & 32.82 & 0.9059 & 10.94 & \underline{33.24} \\
SwinIR + HIR-Diff & 27.88 & 0.5443 & 12.53 & 34.58 & 27.62 & 0.5328 & 13.47 & 43.14 & 28.20 & 0.5456 & 11.84 & 42.82 \\
SwinIR + PWRCTV & 25.43 & 0.5544 & 27.50 & 393.81 & 23.98 & 0.4886 & 29.39 & 662.34 & 22.44 & 0.4233 & 30.17 & 685.00 \\
R-PNN + HIR-Diff & 30.52 & 0.7378 & 10.72 & 25.12 & 29.48 & 0.7004 & 12.24 & 34.86 & 29.48 & 0.7036 & 11.36 & 34.20 \\
R-PNN + PWRCTV & 32.45 & 0.7877 & 13.26 & 26.32 & 30.43 & 0.7176 & 16.23 & 73.68 & 29.56 & 0.6762 & 15.72 & 67.29 \\
HIR-Diff + PLRDiff & 24.58 & 0.4578 & 28.15 & 196.77 & 23.94 & 0.4272 & 28.41 & 80.01 & 25.74 & 0.5280 & 22.85 & 338.78 \\
HIR-Diff + SwinIR & 30.13 & 0.6949 & 10.51 & 26.73 & 29.64 & 0.6803 & 11.85 & 37.69 & 29.51 & 0.6577 & 11.62 & 38.12 \\
HIR-Diff + R-PNN & 33.04 & 0.9030 & 9.29 & 21.81 & 32.67 & 0.8946 & 10.52 & \underline{33.86} & 33.64 & 0.9007 & 9.44 & 33.35 \\
PWRCTV + PLRDiff & 35.00 & 0.8303 & 9.02 & 24.35 & 32.02 & 0.7559 & 15.44 & 234.17 & 31.27 & 0.7379 & 16.33 & 202.54 \\
PWRCTV + SwinIR & 35.65 & 0.8813 & 5.98 & 17.46 & 33.14 & 0.8249 & 11.64 & 219.03 & 32.60 & 0.8146 & 12.07 & 244.81 \\
UHiPandas & \underline{38.65} & \underline{0.9440} & \underline{3.92} & \underline{10.78} & \underline{36.49} & \underline{0.9223} & \underline{8.67} & 43.82 & \underline{36.37} & \underline{0.9218} & \underline{8.70} & 44.13 \\
\midrule
\methodname{} (Ours) & \textbf{39.24} & \textbf{0.9493} & \textbf{3.67} & \textbf{9.48} & \textbf{38.98} & \textbf{0.9469} & \textbf{3.81} & \textbf{9.86} & \textbf{36.91} & \textbf{0.9352} & \textbf{6.13} & \textbf{14.37} \\
\bottomrule
\end{tabular}}
\end{table}

\noindent\textbf{Single-image HSI denoising.}
Tables~\ref{tab:app-qingpu-gaussian}--\ref{tab:app-xiongan-mixed} report six conditions per dataset, each averaged over 85 patches.
BM4D denotes the evaluated GPU approximation.

\begin{table}[!htbp]
\centering\small
\setlength{\tabcolsep}{3pt}
\renewcommand{\arraystretch}{1.08}
\caption{Single-image HSI denoising on Qingpu under Gaussian noise.}
\label{tab:app-qingpu-gaussian}
\resizebox{\linewidth}{!}{%
\begin{tabular}{@{}c*{12}{c}@{}}
\toprule
 & \multicolumn{4}{c}{G30} & \multicolumn{4}{c}{G50} & \multicolumn{4}{c}{G70} \\
\cmidrule(lr){2-5}\cmidrule(lr){6-9}\cmidrule(lr){10-13}
Method & PSNR $\uparrow$ & SSIM $\uparrow$ & SAM $\downarrow$ & ERGAS $\downarrow$ & PSNR $\uparrow$ & SSIM $\uparrow$ & SAM $\downarrow$ & ERGAS $\downarrow$ & PSNR $\uparrow$ & SSIM $\uparrow$ & SAM $\downarrow$ & ERGAS $\downarrow$ \\
\midrule
BM4D & 36.44 & 0.8936 & 2.60 & 6.72 & 33.64 & 0.8194 & 3.65 & 9.29 & 31.77 & 0.7511 & 4.58 & 11.54 \\
HLRTF & 40.70 & 0.9427 & 1.28 & 3.66 & 34.39 & 0.7927 & 2.81 & 8.10 & 29.74 & 0.5845 & 5.11 & 15.13 \\
RCTV & 33.03 & 0.7311 & 4.10 & 10.87 & 27.95 & 0.4866 & 7.36 & 19.66 & 24.79 & 0.3364 & 10.55 & 27.95 \\
DDS2M & 41.57 & 0.9645 & \underline{1.12} & 3.49 & 39.87 & 0.9507 & 1.40 & 4.21 & 38.47 & 0.9361 & 1.68 & 4.95 \\
LRTFR & 38.73 & 0.9235 & 1.39 & 5.11 & 37.11 & 0.9000 & 1.78 & 6.22 & 35.89 & 0.8826 & 2.17 & 7.17 \\
HIR-Diff & 39.31 & 0.9260 & 1.49 & 4.42 & 38.84 & 0.9252 & 1.60 & 4.61 & 37.25 & 0.8958 & 1.95 & 5.51 \\
CF-INR & \underline{42.29} & \underline{0.9686} & 1.14 & \underline{3.25} & \underline{39.94} & \underline{0.9535} & \underline{1.37} & \textbf{4.05} & \underline{38.73} & \underline{0.9403} & \textbf{1.56} & \textbf{4.72} \\
\midrule
\methodname{} (Ours) & \textbf{42.60} & \textbf{0.9714} & \textbf{1.03} & \textbf{3.02} & \textbf{40.05} & \textbf{0.9558} & \textbf{1.35} & \underline{4.09} & \textbf{38.77} & \textbf{0.9413} & \underline{1.60} & \underline{4.73} \\
\bottomrule
\end{tabular}}
\end{table}

\begin{table}[!htbp]
\centering\small
\setlength{\tabcolsep}{3pt}
\renewcommand{\arraystretch}{1.08}
\caption{Single-image HSI denoising on Qingpu under mixed noise.}
\label{tab:app-qingpu-mixed}
\resizebox{\linewidth}{!}{%
\begin{tabular}{@{}c*{12}{c}@{}}
\toprule
 & \multicolumn{4}{c}{M30} & \multicolumn{4}{c}{M50} & \multicolumn{4}{c}{M70} \\
\cmidrule(lr){2-5}\cmidrule(lr){6-9}\cmidrule(lr){10-13}
Method & PSNR $\uparrow$ & SSIM $\uparrow$ & SAM $\downarrow$ & ERGAS $\downarrow$ & PSNR $\uparrow$ & SSIM $\uparrow$ & SAM $\downarrow$ & ERGAS $\downarrow$ & PSNR $\uparrow$ & SSIM $\uparrow$ & SAM $\downarrow$ & ERGAS $\downarrow$ \\
\midrule
BM4D & 22.54 & 0.3971 & 23.26 & 91.66 & 25.36 & 0.5140 & 15.60 & 61.07 & 25.77 & 0.5388 & 13.63 & 53.29 \\
HLRTF & 35.60 & 0.9123 & 27.83 & 115.44 & 30.84 & 0.8113 & 27.54 & 117.36 & 27.25 & 0.6595 & 27.70 & 121.07 \\
RCTV & 36.83 & 0.9183 & 3.88 & 11.74 & 33.64 & 0.8730 & 5.51 & 16.32 & 31.55 & 0.8311 & 7.07 & 20.32 \\
DDS2M & 29.88 & 0.8430 & 13.20 & 53.63 & 29.78 & 0.8552 & 13.12 & 53.49 & 29.70 & 0.8625 & 13.05 & 53.41 \\
LRTFR & 32.70 & 0.8733 & 27.13 & 112.10 & 30.62 & 0.8417 & 26.95 & 114.20 & 29.20 & 0.8176 & 26.94 & 117.34 \\
HIR-Diff & 29.17 & 0.8363 & 13.20 & 55.69 & 29.09 & 0.8295 & 13.21 & 55.69 & 28.91 & 0.8123 & 13.19 & 55.40 \\
CF-INR & \underline{38.71} & \underline{0.9596} & \underline{1.82} & \underline{5.95} & \underline{35.53} & \underline{0.9335} & \underline{2.62} & \underline{8.56} & \underline{33.35} & \underline{0.9086} & \underline{3.42} & \underline{10.99} \\
\midrule
\methodname{} (Ours) & \textbf{39.62} & \textbf{0.9648} & \textbf{1.56} & \textbf{4.35} & \textbf{37.51} & \textbf{0.9389} & \textbf{1.87} & \textbf{5.46} & \textbf{36.50} & \textbf{0.9261} & \textbf{2.24} & \textbf{6.21} \\
\bottomrule
\end{tabular}}
\end{table}

\begin{table}[!htbp]
\centering\small
\setlength{\tabcolsep}{3pt}
\renewcommand{\arraystretch}{1.08}
\caption{Single-image HSI denoising on Xiong'an under Gaussian noise.}
\label{tab:app-xiongan-gaussian}
\resizebox{\linewidth}{!}{%
\begin{tabular}{@{}c*{12}{c}@{}}
\toprule
 & \multicolumn{4}{c}{G30} & \multicolumn{4}{c}{G50} & \multicolumn{4}{c}{G70} \\
\cmidrule(lr){2-5}\cmidrule(lr){6-9}\cmidrule(lr){10-13}
Method & PSNR $\uparrow$ & SSIM $\uparrow$ & SAM $\downarrow$ & ERGAS $\downarrow$ & PSNR $\uparrow$ & SSIM $\uparrow$ & SAM $\downarrow$ & ERGAS $\downarrow$ & PSNR $\uparrow$ & SSIM $\uparrow$ & SAM $\downarrow$ & ERGAS $\downarrow$ \\
\midrule
BM4D & 35.02 & 0.8637 & 2.54 & 5.58 & 32.40 & 0.7786 & 3.49 & 7.57 & 30.66 & 0.7046 & 4.32 & 9.28 \\
HLRTF & \underline{39.91} & \underline{0.9502} & 1.25 & \underline{3.29} & 34.77 & 0.8508 & 2.25 & 5.86 & 30.51 & 0.6857 & 3.84 & 9.86 \\
RCTV & 33.03 & 0.7797 & 3.48 & 7.65 & 27.97 & 0.5534 & 6.25 & 13.88 & 24.79 & 0.3981 & 9.01 & 20.04 \\
DDS2M & 39.75 & 0.9448 & \underline{1.22} & 3.35 & \underline{38.07} & \underline{0.9250} & \underline{1.47} & \underline{4.02} & \underline{36.61} & \underline{0.9001} & \underline{1.73} & \underline{4.73} \\
LRTFR & 35.57 & 0.8309 & 1.61 & 5.70 & 34.18 & 0.7808 & 1.98 & 6.77 & 33.26 & 0.7510 & 2.31 & 7.54 \\
HIR-Diff & 36.40 & 0.8830 & 1.58 & 4.90 & 36.18 & 0.8840 & 1.73 & 4.98 & 35.22 & 0.8652 & 2.01 & 5.55 \\
CF-INR & 39.46 & 0.9446 & 1.47 & 3.50 & 37.67 & 0.9153 & 1.68 & 4.30 & 36.52 & 0.8903 & 1.85 & 4.94 \\
\midrule
\methodname{} (Ours) & \textbf{40.74} & \textbf{0.9632} & \textbf{1.13} & \textbf{2.88} & \textbf{38.66} & \textbf{0.9418} & \textbf{1.42} & \textbf{3.65} & \textbf{36.92} & \textbf{0.9161} & \textbf{1.69} & \textbf{4.49} \\
\bottomrule
\end{tabular}}
\end{table}

\begin{table}[!htbp]
\centering\small
\setlength{\tabcolsep}{3pt}
\renewcommand{\arraystretch}{1.08}
\caption{Single-image HSI denoising on Xiong'an under mixed noise.}
\label{tab:app-xiongan-mixed}
\resizebox{\linewidth}{!}{%
\begin{tabular}{@{}c*{12}{c}@{}}
\toprule
 & \multicolumn{4}{c}{M30} & \multicolumn{4}{c}{M50} & \multicolumn{4}{c}{M70} \\
\cmidrule(lr){2-5}\cmidrule(lr){6-9}\cmidrule(lr){10-13}
Method & PSNR $\uparrow$ & SSIM $\uparrow$ & SAM $\downarrow$ & ERGAS $\downarrow$ & PSNR $\uparrow$ & SSIM $\uparrow$ & SAM $\downarrow$ & ERGAS $\downarrow$ & PSNR $\uparrow$ & SSIM $\uparrow$ & SAM $\downarrow$ & ERGAS $\downarrow$ \\
\midrule
BM4D & 22.38 & 0.3987 & 18.29 & 43.57 & 24.96 & 0.5014 & 12.33 & 28.99 & 25.21 & 0.5095 & 10.85 & 25.93 \\
HLRTF & 33.61 & 0.8941 & 24.44 & 81.55 & 30.12 & 0.8326 & 24.06 & 81.94 & 27.70 & 0.7451 & 24.21 & 83.75 \\
RCTV & 34.86 & 0.8863 & 3.77 & 9.16 & 31.92 & 0.8267 & 5.28 & 12.47 & 30.00 & 0.7754 & 6.75 & 15.83 \\
DDS2M & 29.05 & 0.8231 & 10.05 & 26.26 & 28.93 & 0.8250 & 9.92 & 26.17 & 28.81 & 0.8212 & 9.81 & 26.11 \\
LRTFR & 29.98 & 0.7704 & 23.74 & 79.14 & 28.30 & 0.7195 & 23.51 & 80.07 & 27.23 & 0.6901 & 23.48 & 81.42 \\
HIR-Diff & 28.47 & 0.8351 & 9.83 & 27.49 & 28.41 & 0.8277 & 9.82 & 27.41 & 28.29 & 0.8129 & 9.80 & 27.21 \\
CF-INR & \underline{36.43} & \underline{0.9285} & \underline{2.14} & \underline{5.14} & \underline{33.65} & \underline{0.8762} & \underline{2.95} & \underline{7.30} & \underline{31.82} & \underline{0.8362} & \underline{3.76} & \underline{9.39} \\
\midrule
\methodname{} (Ours) & \textbf{38.00} & \textbf{0.9476} & \textbf{1.62} & \textbf{4.07} & \textbf{37.46} & \textbf{0.9306} & \textbf{1.82} & \textbf{4.40} & \textbf{35.21} & \textbf{0.9035} & \textbf{2.21} & \textbf{5.57} \\
\bottomrule
\end{tabular}}
\end{table}

\noindent\textbf{MRI denoising.}
Table~\ref{tab:app-brainweb-details} reports nine BrainWeb contrast--noise conditions, each averaged over 34 slices.
Table~\ref{tab:app-m4raw-details} reports three M4Raw contrasts, with equal weighting of 25 subjects after averaging 18 slices per subject.
C2S on M4Raw uses the official released weights evaluated on our benchmark. For BrainWeb, we report our own evaluation results for C2S.
DRDD denotes transfer of the official DRDD-AiO5 natural-image pretrained weights.

\begin{table}[!htbp]
\centering\small
\setlength{\tabcolsep}{5pt}
\caption{MRI denoising on BrainWeb across nine contrast--noise conditions.}
\label{tab:app-brainweb-details}
\begin{tabular}{@{}c*{6}{c}@{}}

\multicolumn{7}{@{}c@{}}{\textbf{T1}} \\
\toprule
 & \multicolumn{2}{c}{T1 3\%} & \multicolumn{2}{c}{T1 5\%} & \multicolumn{2}{c}{T1 7\%} \\
\cmidrule(lr){2-3}\cmidrule(lr){4-5}\cmidrule(lr){6-7}
 Method& PSNR $\uparrow$ & SSIM $\uparrow$ & PSNR $\uparrow$ & SSIM $\uparrow$ & PSNR $\uparrow$ & SSIM $\uparrow$ \\
\midrule
Noise2Void & 35.41 & 0.9062 & 34.88 & 0.9425 & 32.97 & 0.8833 \\
Noise2Self & 33.17 & 0.7933 & 33.23 & \underline{0.9600} & 30.81 & 0.7435 \\
Noisier2Noise & 37.49 & 0.9261 & 34.52 & 0.9049 & 31.90 & 0.7828 \\
R2R & 36.20 & 0.8040 & 34.83 & 0.8113 & 32.26 & 0.7534 \\
Noise2Score & 37.48 & 0.8741 & 32.83 & 0.7883 & 29.68 & 0.7370 \\
ZS-N2N & 35.38 & 0.7805 & 31.36 & 0.7240 & 28.70 & 0.6903 \\
C2S & 36.17 & \underline{0.9301} & \underline{34.90} & 0.8402 & \underline{35.05} & \underline{0.8866} \\
DRDD & \underline{38.23} & 0.8314 & 33.85 & 0.7614 & 30.89 & 0.7282 \\
\midrule
\methodname{} (Ours) & \textbf{40.84} & \textbf{0.9824} & \textbf{37.97} & \textbf{0.9737} & \textbf{36.03} & \textbf{0.9636} \\
\bottomrule\addlinespace[5pt]\multicolumn{7}{@{}c@{}}{\textbf{T2}} \\
\toprule
 & \multicolumn{2}{c}{T2 3\%} & \multicolumn{2}{c}{T2 5\%} & \multicolumn{2}{c}{T2 7\%} \\
\cmidrule(lr){2-3}\cmidrule(lr){4-5}\cmidrule(lr){6-7}
 Method & PSNR $\uparrow$ & SSIM $\uparrow$ & PSNR $\uparrow$ & SSIM $\uparrow$ & PSNR $\uparrow$ & SSIM $\uparrow$ \\
\midrule
Noise2Void & 29.46 & 0.8822 & 27.60 & 0.7965 & 26.03 & 0.7304 \\
Noise2Self & 30.26 & 0.8962 & 28.83 & 0.7829 & 28.72 & \underline{0.8620} \\
Noisier2Noise & 32.35 & 0.8728 & 29.29 & 0.8076 & 26.79 & 0.7151 \\
R2R & 33.39 & 0.8196 & 31.10 & 0.8111 & 28.00 & 0.7279 \\
Noise2Score & 32.79 & 0.8268 & 28.42 & 0.7586 & 25.30 & 0.7098 \\
ZS-N2N & 30.89 & 0.7812 & 27.15 & 0.7243 & 24.45 & 0.6807 \\
C2S & \underline{34.39} & \underline{0.9386} & \underline{31.85} & \underline{0.8463} & \underline{29.44} & 0.7749 \\
DRDD & 32.93 & 0.8093 & 28.83 & 0.7595 & 25.98 & 0.7270 \\
\midrule
\methodname{} (Ours) & \textbf{35.36} & \textbf{0.9545} & \textbf{32.46} & \textbf{0.9339} & \textbf{30.54} & \textbf{0.9234} \\
\bottomrule\addlinespace[5pt]\multicolumn{7}{@{}c@{}}{\textbf{PD}} \\
\toprule
 & \multicolumn{2}{c}{PD 3\%} & \multicolumn{2}{c}{PD 5\%} & \multicolumn{2}{c}{PD 7\%} \\
\cmidrule(lr){2-3}\cmidrule(lr){4-5}\cmidrule(lr){6-7}
 Method & PSNR $\uparrow$ & SSIM $\uparrow$ & PSNR $\uparrow$ & SSIM $\uparrow$ & PSNR $\uparrow$ & SSIM $\uparrow$ \\
\midrule
Noise2Void & 31.46 & 0.8547 & 28.73 & 0.8176 & 28.71 & \underline{0.8195} \\
Noise2Self & 30.23 & 0.8943 & 29.14 & 0.7758 & 28.93 & 0.7802 \\
Noisier2Noise & 33.32 & 0.8903 & 30.32 & 0.7897 & 28.00 & 0.7250 \\
R2R & 32.48 & 0.7798 & 30.15 & 0.7413 & 28.24 & 0.7063 \\
Noise2Score & 33.13 & 0.8207 & 28.59 & 0.7394 & 25.58 & 0.6932 \\
ZS-N2N & 31.62 & 0.7686 & 27.70 & 0.7011 & 25.05 & 0.6538 \\
C2S & \underline{35.07} & \underline{0.9174} & \underline{33.14} & \underline{0.8647} & \underline{31.14} & 0.7904 \\
DRDD & 33.93 & 0.8157 & 29.42 & 0.7534 & 26.63 & 0.7148 \\
\midrule
\methodname{} (Ours) & \textbf{36.21} & \textbf{0.9334} & \textbf{33.53} & \textbf{0.9165} & \textbf{31.59} & \textbf{0.8966} \\
\bottomrule
\end{tabular}
\end{table}

\begin{table}[!htbp]
\centering\small
\setlength{\tabcolsep}{3pt}
\renewcommand{\arraystretch}{1.08}
\caption{MRI denoising on M4Raw across three contrasts.}
\label{tab:app-m4raw-details}
\begin{tabular}{@{}c*{6}{c}@{}}
\toprule
 & \multicolumn{2}{c}{T1} & \multicolumn{2}{c}{T2} & \multicolumn{2}{c}{FLAIR} \\
\cmidrule(lr){2-3}\cmidrule(lr){4-5}\cmidrule(lr){6-7}
 Method & PSNR $\uparrow$ & SSIM $\uparrow$ & PSNR $\uparrow$ & SSIM $\uparrow$ & PSNR $\uparrow$ & SSIM $\uparrow$ \\
\midrule
Noise2Void & 31.24 & 0.8671 & 30.45 & 0.8392 & 30.80 & 0.8288 \\
Noise2Self & 29.10 & 0.8537 & 28.17 & 0.8095 & 28.94 & 0.8150 \\
Noisier2Noise & 31.18 & 0.8675 & 30.55 & 0.8435 & 31.17 & 0.8356 \\
R2R & 30.82 & 0.8552 & 30.48 & 0.8329 & 30.97 & 0.8187 \\
Noise2Score & 31.23 & 0.8444 & 30.48 & 0.8139 & 31.00 & 0.7975 \\
ZS-N2N & 31.66 & 0.8922 & 30.96 & 0.8729 & 31.89 & 0.8765 \\
C2S & \underline{32.61} & \underline{0.9152} & \underline{32.30} & \underline{0.8880} & \textbf{33.02} & 0.8792 \\
DRDD & 32.43 & 0.9108 & 31.44 & 0.8770 & 32.60 & \underline{0.8797} \\
\midrule
\methodname{} (Ours) & \textbf{32.69} & \textbf{0.9165} & \textbf{32.31} & \textbf{0.8922} & \underline{32.83} & \textbf{0.8906} \\
\bottomrule
\end{tabular}
\end{table}

\subsection{Additional qualitative results}
\label{app:additional-restoration}
The selected examples use the same display mapping, zoom region, and residual scale across methods within each row.
The left inset shows the zoomed region; the right shows its absolute residual, with brighter colors indicating larger errors.
HSI residuals use displayed RGB channels, BrainWeb uses displayed intensities, and M4Raw uses full-precision intensities.

\begin{figure}[!htbp]
\centering
\includegraphics[page=1,width=\linewidth]{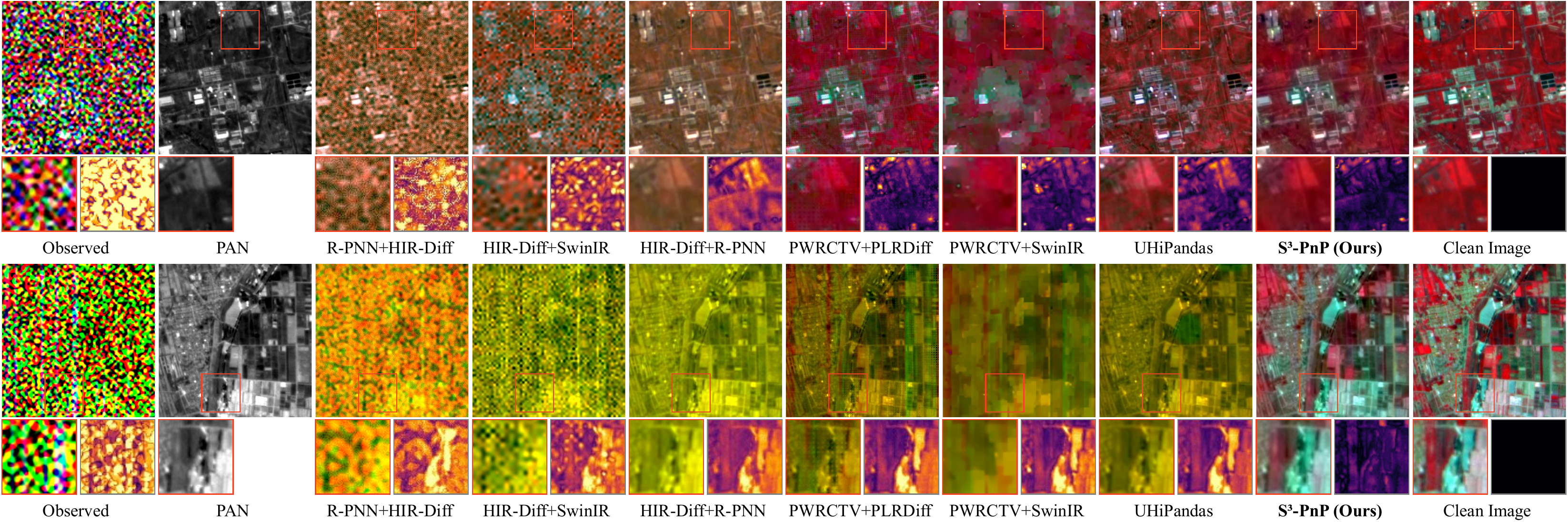}
\caption{Reference-guided HSI restoration on Dongying: G30 (top) and Mix55 (bottom).}
\label{fig:app-dongying}
\end{figure}

\noindent\textbf{Reference-guided HSI restoration.} Figure~\ref{fig:app-dongying} shows Gaussian and mixed-noise examples; PAN is guidance and has no HSI residual.

\begin{figure}[!htbp]
\centering
\includegraphics[page=2,width=\linewidth]{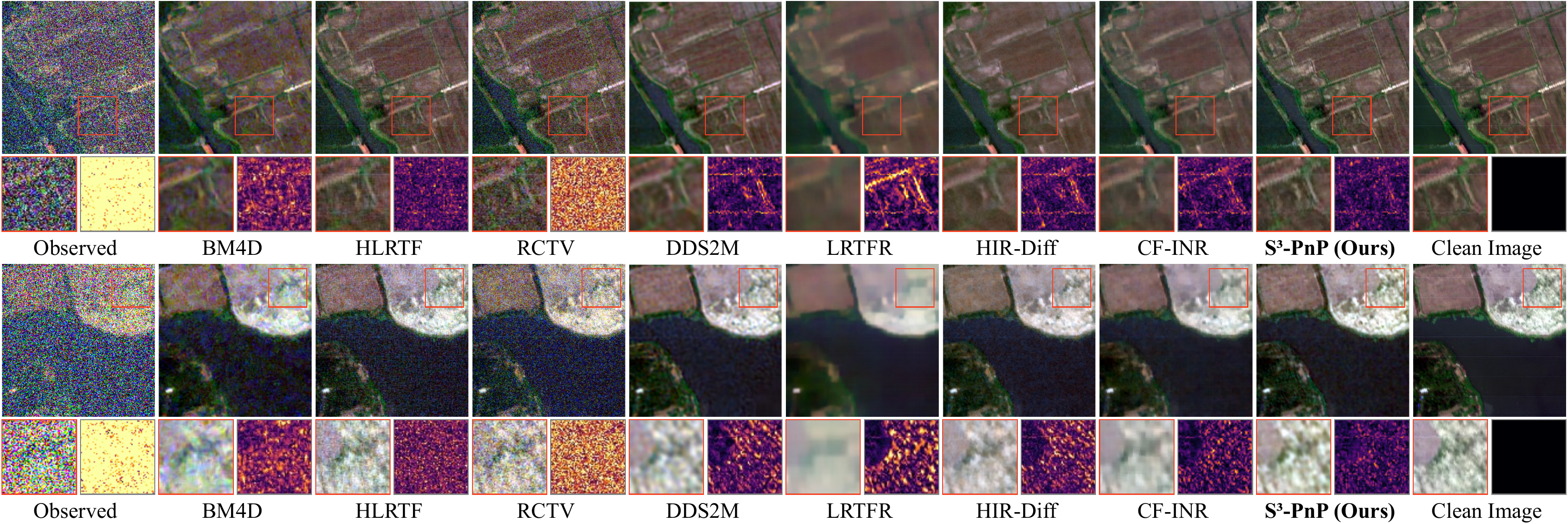}
\caption{Single-image HSI denoising on Qingpu: G30 (top) and G50 (bottom).}
\label{fig:app-qingpu}
\end{figure}

\begin{figure}[!htbp]
\centering
\includegraphics[page=3,width=\linewidth]{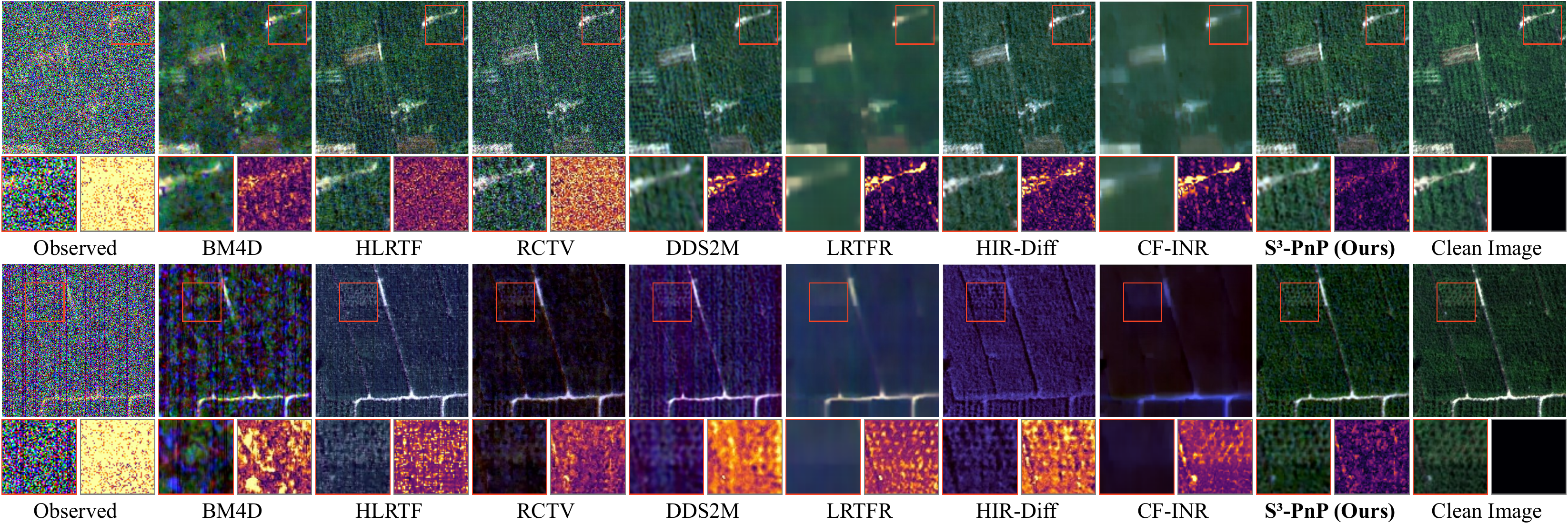}
\caption{Single-image HSI denoising on Xiong'an: G70 (top) and M70 (bottom).}
\label{fig:app-xiongan}
\end{figure}

\noindent\textbf{Single-image HSI denoising.} Figures~\ref{fig:app-qingpu} and~\ref{fig:app-xiongan} show examples from both datasets. 

\begin{figure}[!htbp]
\centering
\includegraphics[page=4,width=\linewidth]{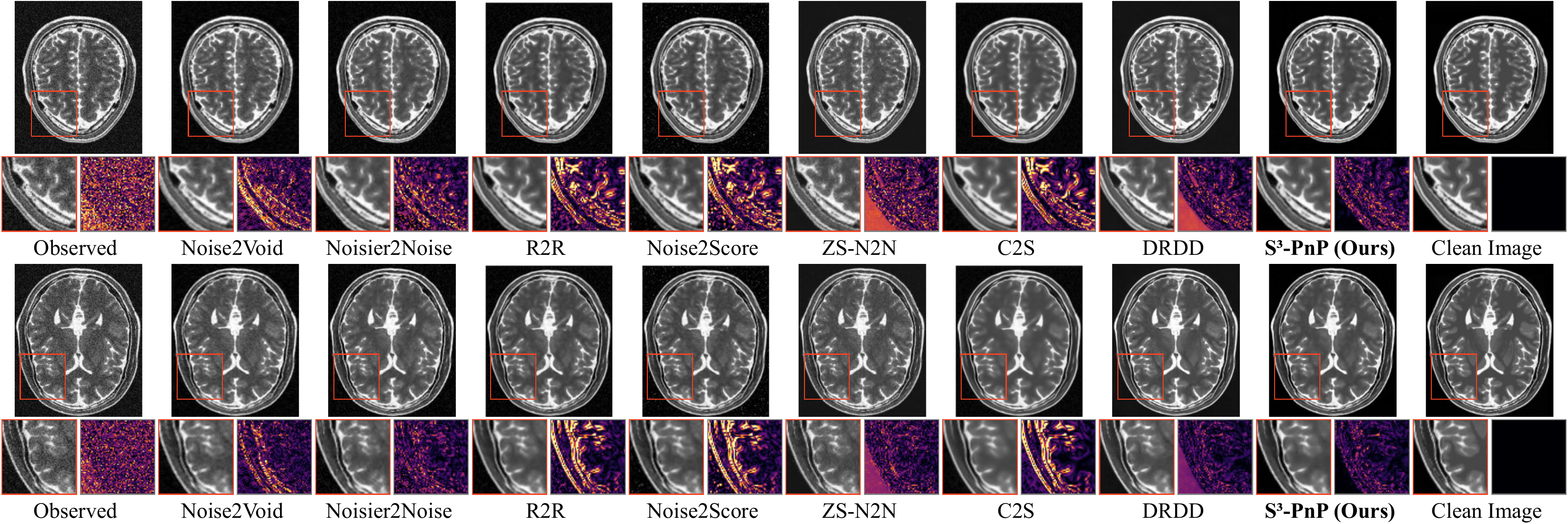}
\caption{MRI denoising on BrainWeb (T2, 7\% noise).}
\label{fig:app-brainweb}
\end{figure}

\begin{figure}[!htbp]
\centering
\includegraphics[page=5,width=\linewidth]{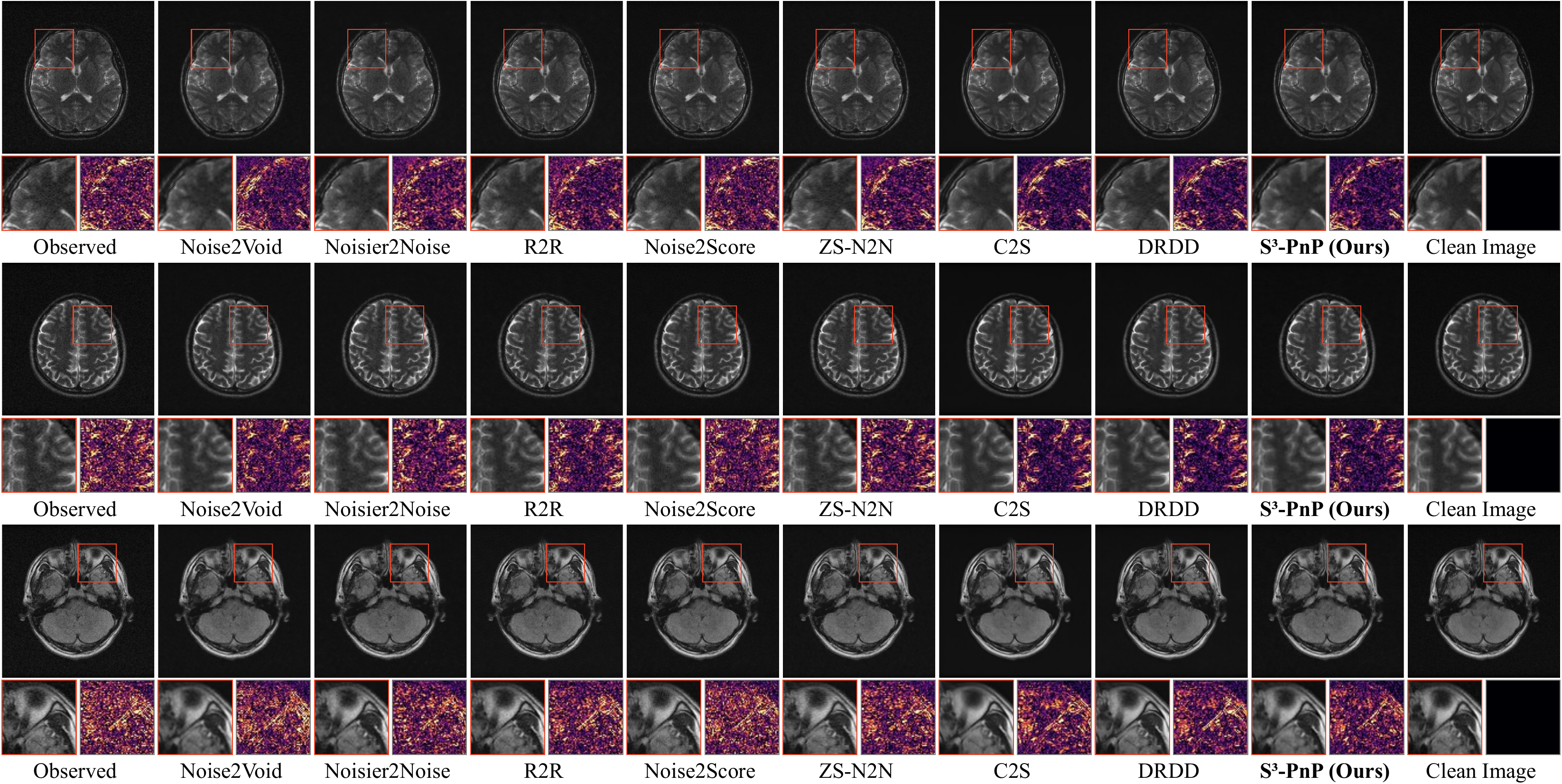}
\caption{MRI denoising on M4Raw: T2, T2, and FLAIR from top to bottom.}
\label{fig:app-m4raw}
\end{figure}

\noindent\textbf{MRI denoising.} Figures~\ref{fig:app-brainweb} and~\ref{fig:app-m4raw} show anatomical details and residuals. The M4Raw reference averages repeated acquisitions, including the input.

\end{document}